\documentclass{article}

\usepackage{arxiv}

\usepackage[utf8]{inputenc} 
\usepackage[T1]{fontenc}    
\usepackage{hyperref}       
\usepackage{url}            
\usepackage{booktabs}       
\usepackage{amsfonts}       
\usepackage{nicefrac}       
\usepackage{microtype}      
\usepackage{lipsum}
\usepackage{subfigure}

\usepackage{amsmath,amsfonts,bm}

\def\eqref#1{equation~\ref{#1}}

\def\1{\bm{1}}

\DeclareMathAlphabet{\mathsfit}{\encodingdefault}{\sfdefault}{m}{sl}
\SetMathAlphabet{\mathsfit}{bold}{\encodingdefault}{\sfdefault}{bx}{n}

\DeclareMathOperator*{\argmax}{arg\,max}

\usepackage{amsmath}
\usepackage[mathscr]{euscript}
\usepackage{amssymb}
\usepackage{tabularx}
\usepackage{textcomp}
\usepackage{gensymb}
\usepackage{multirow}
\usepackage{graphicx}
\usepackage{appendix}

\usepackage{float}
\usepackage{algorithm}
\usepackage[noend]{algpseudocode}

\usepackage{ntheorem}

\usepackage[symbol]{footmisc}

\makeatletter
\newcommand{\printfnsymbol}[1]{%
  \textsuperscript{\@fnsymbol{#1}}%
}
\makeatother

\usepackage[english]{babel}

\usepackage[square,numbers]{natbib}
\title{Learning To Navigate The Synthetically Accessible Chemical Space Using Reinforcement Learning}

\usepackage{authblk}
\author[1]{Sai Krishna Gottipati $^*$}
\author[1]{Boris Sattarov $^*$}
\author[10]{Sufeng Niu}
\author[1,6]{Yashaswi Pathak}
\author[1,9]{Haoran Wei}
\author[2,8]{Shengchao Liu}
\author[1]{Karam J. Thomas}
\author[8]{Simon Blackburn}
\author[7]{Connor W. Coley}
\author[5,8,11]{Jian Tang}
\author[3,5,8]{Sarath Chandar}
\author[2,4,5,8]{Yoshua Bengio}

\affil[1]{99andBeyond}
\affil[2]{University of Montreal}
\affil[3]{Ecole Polytechnique Montréal}
\affil[4]{CIFAR Senior Fellow}
\affil[5]{Canada CIFAR AI Chair}
\affil[6]{Center for Computational Natural Sciences and Bioinformatics, IIIT Hyderabad}
\affil[7]{Department of Chemical Engineering, Massachusets Institute of Technology}
\affil[8]{Mila - Quebec AI Institute}
\affil[9]{University of Delaware}
\affil[10]{Clemson University, South Carolina}
\affil[11]{HEC Montréal}

\begin{document}
\def\thefootnote{*}\footnotetext{Equal contribution}
\maketitle

\begin{abstract}
Over the last decade, there has been significant progress in the field of machine learning for de novo drug design, particularly in deep generative models. However, current generative approaches exhibit a significant challenge as they do not ensure that
the proposed molecular structures can be feasibly synthesized nor do they provide the synthesis routes of the proposed small molecules, thereby seriously limiting their practical applicability. In this work, we propose a novel forward synthesis framework powered by reinforcement learning (RL) for de novo drug design, Policy Gradient for Forward Synthesis (PGFS), that addresses this challenge by embedding the concept of synthetic accessibility directly into the de novo drug design system. In this setup, the agent learns to navigate through the immense synthetically accessible chemical space by subjecting commercially available small molecule building blocks to valid chemical reactions at every time step of the iterative virtual multi-step synthesis process. The proposed environment for drug discovery provides a highly challenging test-bed for RL algorithms owing to the large state space and high-dimensional continuous action space with hierarchical actions. PGFS achieves state-of-the-art performance in generating structures with high QED and penalized clogP. Moreover, we validate PGFS in an in-silico proof-of-concept associated with three HIV targets. Finally, we describe how the end-to-end training conceptualized in this study represents an important paradigm in radically expanding the synthesizable chemical space and automating the drug discovery process.
\end{abstract}


\section{Introduction}

In the last decade, the role of machine learning and artificial intelligence techniques in chemical sciences and drug discovery has substantially increased (\citet{schneider_automating_2018,nature_ml_molecular_science,dlforcompchemistry}). Deep generative models  such as GANs and VAEs have emerged as promising new techniques to design novel molecules with desirable properties (\citet{sanchez-lengeling_inverse_2018,assouel2018defactor,moleculegenerationreview}). Generative models using either string-based (e.g., \citet{segler_generating_2017}) or graph-based representations (e.g., \citet{jtvae}) are able to output chemically valid molecules in a manner that can be biased towards properties like drug-likeness.

\begin{figure}
    \begin{center}
    \includegraphics[width=1.0\columnwidth]{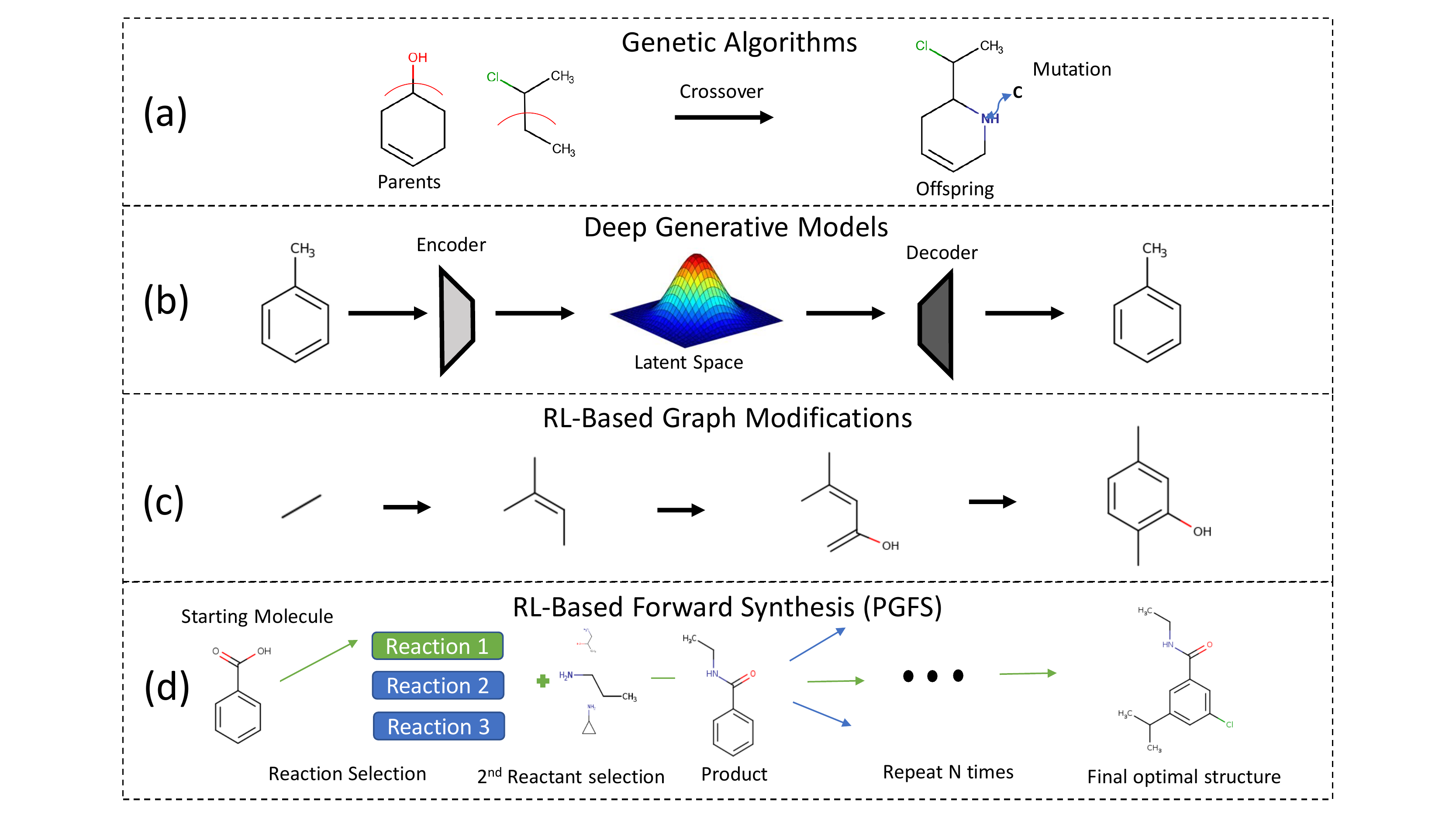}
    \end{center}
    \vspace{-3mm}
    \caption{
    Illustrative comparison of de novo drug design methodologies including: (a) genetic (evolutionary) algorithms (\citet{brown2004graph, jensen2019graph}); (b) deep generative models (~\citet{simonovsky2018graphvae,gomez2018automatic,winter2019efficient,jtvae,popova2018deep,olivecrona2017molecular}); (c) RL-based graph modifications (~\citet{gcpn,MolDQN}); and (d) RL-based forward synthesis as proposed in our methodology Policy Gradient for Forward Synthesis (PGFS).
    }
    \label{fig:intro_toc}
\end{figure}

However, the majority of de novo drug design methodologies do not explicitly account for synthetic feasibility, and thus cannot ensure whether the generated molecules can be produced in the physical world. Synthetic complexity scores (\citet{ertl2009estimation, coley2018scscore}) can be introduced into the scoring function to complement generative models. However, like any other data-driven predictive model, these heuristics are prone to exploitation by the generator, i.e, certain generated molecules with high accessibility scores will still be impossible or challenging to produce (\citet{connor_new}). Even though there is great work that has been done in the field of computer aided synthesis planning (\citet{szymkuc_computer-assisted_2016,nature-planningsynthesis,coley_machine_2018,coley_robotic_2019}), relying on these programs creates a disjoint search pipeline that necessitates a separate algorithm for molecule generation and never guarantees that the generative model learns anything about synthesizability. 

Directly embedding synthetic knowledge into de novo drug design would allow us to constrain the search to synthetically-accessible routes and theoretically guarantee that any molecule proposed by the algorithm can be easily produced. 
To accomplish this, we present a forward synthesis model powered by reinforcement learning (RL) entitled Policy Gradient for Forward Synthesis (PGFS) that treats the generation of a molecular structure as a sequential decision process of selecting reactant molecules and reaction transformations in a linear synthetic sequence. The agent learns to select the best set of reactants and reactions to maximize the task-specific desired properties of the product molecule, i.e., where the choice of reactants
is considered an action, and a product molecule is a state of the system obtained through a trajectory
composed of the chosen chemical reactions. The primary contribution of this work is the development of a RL framework able to cope with the vast discrete action space of multi-step virtual chemical synthesis and bias molecular generation towards chemical structures that maximize a black-box objective function, generating a full synthetic route in the process. We define the problem of de novo drug design via forward synthesis as a Markov decision process in chemical reaction space, and we propose to search in a continuous action space using a relevant feature space for reactants rather than a discrete space to facilitate the learning of the agent. Training is guided by rewards which correspond to the predicted properties of the resulting molecule relative to the desired properties. We show that our algorithm achieves state-of-the-art performance on standard metrics like QED and penalized clogP. Furthermore, as a proof-of-concept, our algorithm generated molecules with higher predicted activity against three HIV-related biological targets relative to existing benchmarks. The HIV targets activity datasets used, predictive QSAR models and prediction scripts can be found at this url: \url{https://github.com/99andBeyond/Apollo1060}

\section{Related Work}

To highlight the improvements we are proposing in this work, we focus the discussion on de novo drug design methodologies that can perform single- and multi-objective optimization of chemical structures.

\subsection{Genetic Algorithms}

Genetic algorithms (GA) have been used for many decades to generate and optimize novel chemical structures. They represent one of the most straightforward and simple approaches for de novo drug design and can perform on-par with complex generative models across popular benchmark tasks(\citet{jensen2019graph}). The majority of published GA approaches (\citet{brown2004graph, jensen2019graph}) use graph-based representations of the molecule and apply specific graph sub-fragments crossover operations to produce offsprings followed by mutation operations in the form of random atom, fragment and bond type replacements. More recently, string-based representations of molecules were also proposed in the GA optimization setting (\citet{krenn2019selfies, nigam2019augmenting}). Existing implementations of GA for de novo generation can only account for synthetic feasibility through the introduction of a heuristic scoring functions (\citet{ertl2009estimation, coley2018scscore}) as part of the reward function. As a result, they need a separate model for retrosynthesis or manual evaluation by an expert upon identifying a structure with desired properties.

\subsection{Deep Generative Models}
Many recent studies highlight applications of deep generative systems in multi-objective optimization of chemical structures (\citet{gomez2018automatic,winter2019efficient}). Other recent publications describe improvements in learning by utilizing RL (\citet{olivecrona2017molecular, popova2018deep,guimaraes_objective-reinforced_2017}). While these approaches have provided valuable techniques for optimizing various types of molecular properties in single- and multi-objective settings, they exhibit the same challenges in synthetic feasibility as genetic algorithms.

\subsection{RL-Based Graph Modification Models}
\citet{you2018graph} and \citet{MolDQN} recently proposed reinforcement learning based algorithms to iteratively modify a molecule by adding and removing atoms, bonds or molecular subgraphs. In such setups, the constructed molecule $M_t$, represents the state at time step $t$. The state at time step $0$ can be a single atom like carbon or it can be completely null. The agent is trained to pick actions that would optimize the properties of the generated molecules. While these methods have achieved promising results, they do not guarantee synthetic feasibility.    

\subsection{Forward Synthesis Models}
The generation of molecules using forward synthesis is the most straightforward way to deal with the problem of synthetic accessibility. Generalized reaction transformations define how one molecular subgraph can be produced from another and can be encoded by expert chemists (\citet{hartenfeller2012dogs,szymkuc_computer-assisted_2016}) or algorithmically extracted from reaction data (\citet{law_route_2009}). Libraries of these ``templates'' can be used to enumerate hypothetical product molecules accessible from libraries of available starting materials. In fact, de novo drug design via forward synthesis isn't a new concept, and has been used for decades to generate chemical libraries for virtual screening (\citet{walters_virtual_2018}). Templates can be used in a goal-directed optimization setting without relying on complete enumeration. \citet{vinkers2003synopsis} describe an iterative evolutionary optimization approach called SYNOPSIS to produce chemical structures with optimal properties using reaction-based transformations. 
 \citet{patel2009knowledge} explored the enumeration and optimization of structures by taking advantage of the reaction vectors concept. More recently, many approaches focused on reaction-based enumeration of analogs of known drugs and lead compounds have been proposed (\citet{hartenfeller2012dogs, button2019automated}). Although promising results were reported when using a reaction-based enumeration approach that was followed by an active learning module (\citet{konze2019reaction}), mere enumeration severely limits the capacity of the model to explore the chemical space efficiently. 

Recently, \citet{bradshaw2019} and \citet{korovina} have proposed approaches to de novo drug design that use reaction prediction algorithms to constrain the search to synthetically-accessible structures. \citet{bradshaw2019} use a variational auto-encoder to embed reactant structures and optimize the molecular properties of the resulting product from the \emph{single-step} reaction by biasing reactant selection. \citet{korovina} propose an algorithmically simpler approach, whereby random selection of reactants and conditions are used to stochastically generate candidate structures, and then subject the structures to property evaluation. This workflow produces molecules through multi-step chemical synthesis, but the selection of reactants cannot be biased towards the optimization objective. We combine the unique strengths of both frameworks (biased generation and multi-step capabilities) in our approach; in doing so, we make use of a novel RL framework.

\subsection{Benchmarking De Novo Drug Design}

It is difficult to properly evaluate approaches for de novo drug design without conducting the actual synthesis of the proposed compounds and evaluating their properties in laboratory experiments. Yet, several simple benchmarks have been adopted in recent publications. Metrics like the Frechenet ChemNet distance (\citet{preuer2018frechet}) aim to measure the similarity of the distributions of the generated structures relative to the training set. Objective-directed benchmarks evaluate the ability to conduct efficient single- and multi-objective optimization for the proposed structures. The most widely used objective functions are QED (\citet{bickerton2012quantifying}), a quantitative estimate of drug-likeness, and penalized clogP as defined by \citet{you2018graph}, an estimate of the octanol-water partition coefficient that penalizes large aliphatic cycles and molecules with large synthetic accessibility scores (\citet{ertl2009estimation}). While these metrics enable the comparison of systems with respect to their ability to optimize simple reward functions associated with the proposed structures, they bear little resemblance to what would be used in a real drug discovery project. 
Recently, two efforts in creating benchmarking platforms have been described in the corresponding publications: MOSES (\citet{polykovskiy2018molecular}) and GuacaMol (\citet{brown2019guacamol}). While MOSES focuses on the distribution of properties of the generated structures, GuacaMol aims to establish a list of goal-directed drug de novo design benchmarks based on the similarity to a particular compound, compound rediscovery and search for active compounds containing different core structures (scaffold hopping). In a recent review describing the current state of the field (\citet{coley2019autonomous}) of autonomous discovery, the authors state that the community needs to focus on proposing benchmarks that will better incorporate the complexity of the real-world drug discovery process such as ligand and structure based modeling, docking, Free Energy Perturbation Calculations (FEP), etc.

\section{Methods}

\subsection{Reinforcement Learning}

To explore the large chemical space efficiently and maintain the ability to generate diverse compounds, we
propose to consider a molecule as a sequence of unimolecular or bimolecular reactions applied to an initial molecule. PGFS learns to select the best set of commercially available reactants and reaction templates that maximize the rewards associated with the properties of the product molecule. This guarantees that the only molecules being considered are synthesizable and also provides the recipe for synthesis. The state of the system at each step corresponds to a product molecule and the rewards are computed according to the properties of the product. Furthermore,
our method decomposes actions of synthetic steps in two sub-actions. A reaction template is first selected and is followed by the selection of a reactant compatible with it. This hierarchical decomposition considerably reduces the size of the action space
in each of the time steps in contrast to simultaneously picking a reactant and reaction type.   

However, this formulation still poses challenges for current state-of-the-art RL algorithms like PPO (\citet{PPO}) and ACKTR (\citet{ACKTR}) owing to the large action space. In fact, there are tens of thousands of possible reactants for each given molecule and reaction template. As a result, we propose to adapt algorithms corresponding to continuous action spaces and map continuous embeddings
to discrete molecular structures by looking up the nearest molecules in this representation space via
a k-nearest neighbor (k-NN) algorithm. Deterministic policy gradient (\citet{DPG}) is one of the first RL algorithms for continous action space. Deep deterministic policy gradient (DDPG) (\citet{DDPG}), Distributed distributional DDPG (D4PG) (\citet{D4PG}) and Twin delayed DDPG (TD3) (\citet{TD3}) constitute the consequent improvements done over DPG. Soft actor critic (SAC, \citet{SAC}) also deals with continuous action spaces with entropy regularization. In this work, we leverage a TD3 algorithm along with the k-NN approach from \citet{LargeDiscreteActionSpaces}. There are three key differences with this work: (1) our actor module includes two learnable networks (instead of just one) to compute two levels of actions; (2) we do not use a critic network in the forward propagation, and include the k-NN computation as part of the environment. Thus, the continuous output of the actor module reflects the true actions--not a proto-action to be discretized to obtain the actual action; and (3) we leverage the TD3 algorithm which has been shown to be better than DPG (used in \citet{LargeDiscreteActionSpaces}) on several RL tasks.

\subsection{Overview}

The pipeline is setup in such a way that at every time step $t$, a reactant $R^{(2)}_t$ is selected to react with the existing molecule $R^{(1)}_t$ to yield the product  $R^{(1)}_{t+1}$ which is the molecule for the next time step. $R^{(1)}_t$ is considered as the  current state $s_t$ and our agent chooses an action $a_t$ that is further used in computing $R^{(2)}_t$. The product $R^{(1)}_{t+1}$ (which is considered as the next state $s_{t+1}$) is determined by the environment based on the two reactants ($R^{(1)}_t$ and $R^{(2)}_t$). At the very initial time step, we randomly sample the initial molecule $R^{(1)}_0$ from the list of all commercially available reactants. To overcome the limitation of large discrete action space where there are over a hundred thousand possible second reactants, we introduce an intermediate action which reduces the space of reactants considered by choosing a reaction template. Reaction templates, encoded in the SMARTS (\citet{james2000smarts}) language, define allowable chemical transformations according to subgraph matching rules. They can be applied deterministically to sets of reactant molecules to propose hypothetical product molecules using cheminformatics tools like RDKit (\citet{rdkit}). One of the reactants is the state $s_t$ while the other reactant is later selected. Since the required substructure of $R^{(2)}_t$ that can participate in the reaction and of the state $s_t$ is determined by the choice of the reaction template, the action space comprising the space of all $R^{(2)}$s becomes constrained to those reactants which contain this particular substructure. We also enforce the additional constraint of having this substructure present only once in the structure. 
Pairs of reactants associated with different templates that can participate in several different reactions are also forbidden. When multiple products are still possible, one of them is selected randomly. Even with the previous constraints, there can be tens of thousands of reactants at each step, and thus represents a challenge for traditional RL algorithms. Thus, we formulate a novel Markov Decision Process (MDP) involving a continuous action space.

The agent comprises three learnable networks $f$, $\pi$ and $Q$. In terms of the actor-critic framework, our actor module $\Pi$ comprises $f$ and $\pi$ networks and the critic is composed of the $Q$ network that estimates the Q-value of the state-action pair. 
At any time step $t$, the input to the actor module is the state $s_t$ ($R^{(1)}_t$) and the output is the action $a_t$ which is a tensor defined in the feature representation space of all initial reactants $R^{(2)}$. The $f$ network predicts the best reaction template $T_t$ given the current state $s_t$ ($R^{(1)}_t$). Using the best reaction template $T_t$ and ($R^{(1)}_t$) as inputs, the $\pi$ network computes the action $a_t$.
The environment takes the state $s_t$, best reaction template $T_t$, and action $a_t$ as inputs and computes the reward $r_t$, next state $s_{t+1}$ and a boolean to determine whether the episode has ended. It first chooses $k$ reactants from the set $R^{(2)}$ corresponding to the $k$-closest embeddings to the action $a$ using the k nearest neighbours technique in which we pre-compute feature representations for all reactants. Each of these $k$ actions are then passed through a reaction predictor to obtain the corresponding $k$ products. The rewards associated with the products are computed using a scoring function. The reward and product corresponding to the maximum reward are returned. The state $s_t$, best template $T_t$, action $a_t$, next state $s_{t+1}$, reward $r_t$ are stored in the replay memory buffer.
The episode terminates when either the maximum number of reaction steps is reached or when the next state has no valid templates. In our experiments, we have 15 unimolecular and 82 bimolecular reaction templates. The unimolecular templates do not require selection of an $R^{(2)}$, and hence for such cases we directly obtain the product using $R^{(1)}_t$ and the selected $T_t$.  

\begin{figure}
    \centering
    \includegraphics[width=0.8\columnwidth]{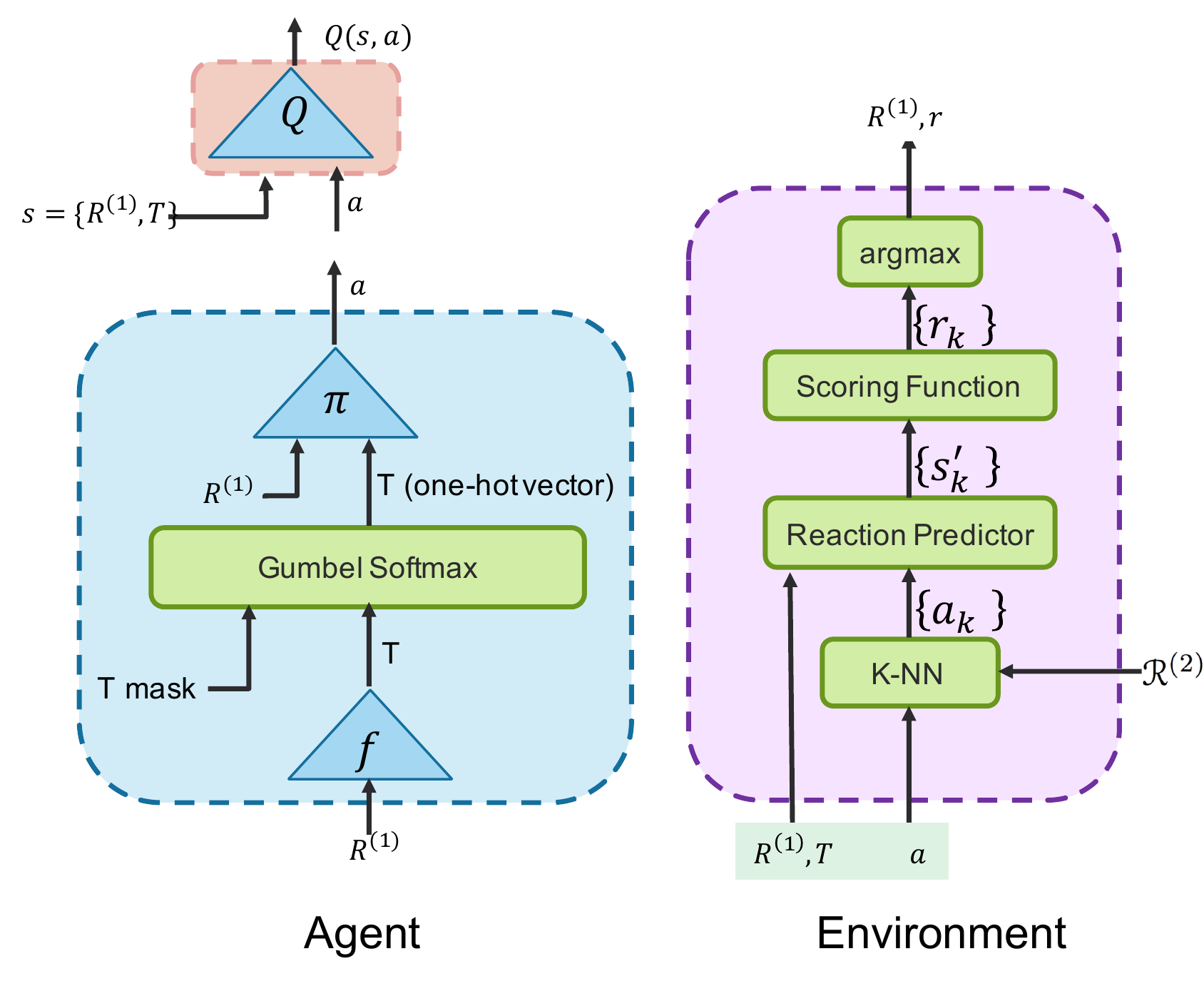}
    \vspace{-0.6cm}
    \caption{PGFS environment and agent. 
    The environment takes in the state $s$ ($R^{(1)}$), the reaction template $T$, the action $a$ (a tensor in the space of feature representations of reactants) and outputs the next state $s'$ ($R^{(1)}$ for next time step) and reward $r$. $\{a_k\}$ is the set of top-k reactants closest to the action $a$, obtained using the k-nearest neighbours algorithm. The reaction predictor computes k products (next states) corresponding to the k reactants when they react with $R^{(1)}$. The scoring function computes k rewards corresponding to the k next states. Finally, the next state corresponding to the maximum reward is chosen. The agent is composed of actor and critic modules. The actor predicts the action $a$ given the state input $R^{(1)}$ and the critic evaluates this action.
    }
    \label{fig:block-diagram}
\end{figure}

During initial phases of the training, it is important to note that the template chosen by the $f$ network might be invalid. To overcome this issue and to ensure the gradient propagation through the $f$ network, we first multiply the template $T$ with the template mask $T_{mask}$ and then use Gumbel softmax to obtain the best template:
$$ T = T \odot T_{mask} $$ 
$$ T = GumbelSoftmax(T, \tau) $$
where, $\tau$ is the temperature parameter that is slowly annealed from 1.0 to 0.1.

\subsubsection{Training Paradigm}
The learning agent can be trained using any policy gradient algorithm applicable for continuous action spaces. Thus, we call our algorithm "Policy Gradient for Forward Synthesis (PGFS)". DDPG (\citet{DDPG}) is one of the first deep RL algorithms for continuous action spaces. After sampling a random minibatch of $N$ transitions from the buffer, the actor and critic modules are updated as follows: 
The critic ($Q$-network) is updated using the one-step TD update rule as: $$y_i = r_i + \gamma Q'(s_{i+1}, \Pi'(s_{i+1})) $$
where, $Q'$ and $\Pi'$ are the target critic and actor networks respectively, i.e, they are a copy of the original networks but they do not update their parameters during gradient updates. $y_i$ is the one-step TD target, $r_i$ is the immediate reward and $s_i$ constitutes the state at the time step $t$. $s_{i+1}$ forms the state at next time step. The critic loss is then:
$$ L = \frac{1}{N} \sum_i (y_i - Q(s_i, a_i))^2 $$
and the parameters of the $Q$ network are updated via back propagation of the critic loss. The goal of the actor module is to maximize the overall return (weighted average of future rewards) achieved over the given initial distribution of states by following the actions determined by the actor module. The $Q$ network can be seen as an approximation to this overall return. Thus, the actor should predict actions that maximize the $Q(s,a)$ values predicted by $Q$ network i.e, $  \max Q(s, \Pi(s)) $, or $\min -Q(s, \Pi(s))$. Thus, $-Q(s,\Pi(s))$ constitutes the actor loss. Consequently, the parameters of the actor module (of $f$ and $\pi$ networks) are updated towards reducing that loss. 

However, the convergence of returns observed is slow because of the reasons highlighted by \citet{TD3}. Accordingly, we use the approach from \cite{TD3} for faster training. 

Firstly, we smooth the target policy (akin to regularization strategy) by adding a small amount of clipped random noises to the action. 
$$ \Tilde{a} = a + \epsilon;~~~ \epsilon \sim \text{clip}(N(0, \Bar{\sigma}), -c, c) $$ 

We use a double Q-learning strategy comprising two critics, but only consider the minimum of two critics for computing the TD target:
$$ y = r + \gamma \min_{i=1,2} Q_i(s',\Pi(s')) $$

Further, we make delayed updates (typically, once every two critic updates) to the actor module and target networks. 

To speed up the convergence of the $f$ network, we also minimize the cross entropy between the output of the $f$ network and the corresponding template $T$ obtained for the reactant $R^{(1)}$.

\begin{algorithm}
\caption{PGFS}
\begin{algorithmic}[1]
\label{alg:CaQL}

\Procedure{Actor}{$R^{(1)}$}
\State $T \gets f(R^{(1)})$
\State $T \gets T \odot T_{mask}$
\State $T \gets GumbelSoftmax(T, \tau)$
\State $a \gets \pi(R^{(1)}, T)$
\State return $T, a$
\EndProcedure

\Procedure{Critic}{$R^{(1)}$, $T$, $a$}
\State return $Q(R^{(1)}, T, a)$
\EndProcedure

\Procedure{env.step}{$R^{(1)}, T, R^{(2)}$}
\State $\mathscr{R}^{(2)} \gets$ GetValidReactants$(T) $
\State $\mathscr{A} \gets$ kNN$(a, \mathscr{R}^{(2)}) $
\State $\mathscr{R}^{(1)}_{t+1} \gets$ ForwardReaction$(R^{(1)}, T, \mathscr{A})$
\State $\mathscr{R}ewards \gets$ ScoringFunction$(\mathscr{R}^{(1)}_{t+1})$
\State $r_t, R^{(1)}_{t+1}, done \gets \argmax \mathscr{R}ewards$
\State return $R^{(1)}_{t+1}, r_t$, done 
\EndProcedure

\Procedure{backward}{buffer minibatch}

\State $T_{i+1}, a_{i+1} \gets$ Actor-target$(R^{(1)}_{i+1}$)
\State $y_i \gets r_i + \gamma$ $\min_{j=1,2}$ Critic-target$(\{ R^{(1)}_{i+1}, T_{i+1}\}, a_{i+1})$ 
\State min $L(\theta^{Q}) = \frac{1}{N} \sum_i |y_i -$ Critic$(\{R^{(1)}_i, T_i \}, a_i)|^2 $
\State $\min L(\theta^{f, \pi}) = -\sum_i Critic(R^{(1)}_i, Actor(R^{(1)}_i)) $
\State $\min L(\theta^{f}) = -\sum_i (T^{(1)}_{i}, log(f(R^{(1)}_{i}))) $

\EndProcedure

\Procedure{main}{$f$, $\pi$, $Q$}

\For{episode = 1, M}
\State sample $R^{(1)}_0$
\For{t = 0, N}

\State $T_t, a_t \gets $ Actor($R^{(1)}_t$)
\State $R^{(1)}_{t+1}, r_t$, done $\gets$ env.step$(R^{(1)}_t, T_t, a_t)$

\State store $(R^{(1)}_t, T_t, a_t, R^{(1)}_{t+1}, r_t$, done) in buffer
\State sample a random minibatch from buffer
\State Backward(minibatch)

\EndFor
\EndFor
\EndProcedure
\end{algorithmic}
\end{algorithm}

\section{Experiments}

\subsection{Predictive Modeling}
To test the applicability of PGFS in an in-silico proof-of-concept for de novo drug design, we develop predictive models against three biological targets related to the human immunodeficiency virus (HIV) - as scoring functions. The biological activity data available in the public domain allowed us to develop ligand-based machine learning models using the concept of quantitative structure-activity relationship modeling (QSAR). 

\paragraph{HIV Targets}

i) The first target in this study, C-C chemokine receptor type 5 (CCR5), is a receptor located on the surface of the host immune cells. Along with C-X-C chemokine receptor type 4 (CXCR4), this receptor is used by HIV to recognize target cells. Hence, antagonists of this receptor allows HIV entry inhibition (\citet{arts2012hiv}). 

ii)
The second target is HIV integrase that catalyzes HIV viral DNA processing and strand transfer. Inhibitors of that enzyme target the strand transfer reaction, thus allowing for HIV integration inhibition. 

iii)
The last selected target is HIV Reverse transcriptase (HIV-RT) which was the first enzyme used as biological target in antiretroviral drug discovery. It is an enzyme with multiple functions that are necessary to convert the single strand of the viral RNA to a double stranded DNA. 

\paragraph{Quantitative Structure Activity Relationships}
The goals of QSAR studies is to discover functional relationship between the structure of the chemical compound and its activity relative to a biological target of interest (\citet{cherkasov2014qsar}). This is achieved by training a supervised machine learning model based on chemical descriptors (\citet{tuppurainen1999frontier}) as feature inputs to predict a biological response value of the compound referring to the half maximal inhibitory concentration, IC50. Widely accepted  guidelines for building QSAR models (\citet{tropsha2010best, cherkasov2014qsar, muratov2020qsar}) were developed describing training data curation, testing models performance, usage of the Applicability Domain (AD) and more. We trained our QSAR models to predict the compounds' pIC$_{50}$ (-log$_{10}$IC$_{50}$ where the IC$_{50}$ is the molar concentration of a compound that produces a half-maximum inhibitory response) values associated with three HIV-related targets reported in the ChEMBL database. The description of the data curation, QSAR training procedures and definition of AD can be found in Section-\ref{sec:appendix-c} of the Appendix. The predictive performance of the developed models can be found in Table-\ref{tab:qsar_performance_table}. The predictive performance metrics were calculated using five-fold cross validation repeated five times. In this work, QSAR models were used as a more complex design objective than other scoring functions (QED, penalized clogP). Proper application of such models in drug discovery will require more rigorous data curation, stricter AD formulation, and more thorough performance testing using scaffold- and/or time-split tests depending on the target of interest.

\begin{table}

\centering

\caption{Cross-validation performance values for the trained QSAR models. Presented metrics and values: $R^2$ - coefficient of determination, $R_{adj}^2$ - adjusted coefficient of determination (\citet{srivastava1995coefficient}), \emph{MAE} - mean absolute error, \emph{Range} - range of the response values (pIC50) in the dataset. Values in the \emph{average} column were calculated by averaging the performance of the 25 models (five-fold cross validation repeated five times) on their unique random validation sets. Values in the \emph{aggregated} column were calculated by combining all out-of-fold predictions: Each compound instance was predicted as the average prediction of only five models that were built while the instance was not present in the training set of these models during the five-fold cross validation repeated five times. These prediction values were compared with the ground truth. The formulas for calculating the metrics as well as measured vs. predicted plots are presented in the Appendix Section - \ref{sec:appendix-c}}

\begin{tabular}{|c|c|c|c|c|c|c|c|}
\hline
Dataset & \multicolumn{2}{c|}{$R_{adj}^2$} & \multicolumn{2}{c|}{$R^2$} & \multicolumn{2}{c|}{MAE} & Range \\\hline
        & aggregated       & average      & aggregated    & average   & aggregated   & average  &       \\\hline
CCR5    & 0.72             & $0.64\pm{0.03}$ &0.72& $0.69\pm{0.03}$&0.51  & $0.54\pm{0.02}$  &4.04-10.30 \\\hline
HIV-Int & 0.68 & $0.45\pm{0.07}$& 0.69  & $0.65\pm{0.04}$& 0.45  & $0.48\pm{0.03}$ & 4.00-8.15\\\hline
HIV-RT  & 0.53 & $0.40\pm{0.06}$& 0.55  & $0.52\pm{0.05}$& 0.51  & $0.53\pm{0.03}$ & 4.00-8.66 \\\hline     
\end{tabular}
\label{tab:qsar_performance_table}
\end{table}

\subsection{Data And Representations}


\paragraph{ECFP4-Like Morgan Fingerprints}
We utilize Morgan circular molecular fingerprint bit vector of size 1024 and radius 2 as implemented in RDKit (\citet{rdkit}) with default invariants that use connectivity information similar to those used for the ECFP fingerprints (\citet{rogers2010extended}). Generally, Morgan fingerprints utilize the graph topology, and thus can be viewed as a learning-free graph representation. Moreover, some recent studies (\citet{NIPS2019_9054}) demonstrate its performance is competitive to the state-of-the-art Graph Neural Network.
\paragraph{MACCS Public Keys}
We leveraged 166 public MACCS keys as implemented in RDKit. MACCS keys constitute a very simple binary feature vector where each bin corresponds to the presence (1) or to the absence (0) of the pre-defined molecular sub-fragment. 
\paragraph{Molecular Descriptors Set (MolDSet)}
The set of normalized molecular continuous descriptors are selected from the 199 descriptors available in RDKit (\citet{rdkit}). The selection was done according to the predicted QSAR results against the three HIV targets reported in this study. The set consists of descriptors that were selected during the feature selection process in every one of 25 models for each target. The resulting set of 35 features consists of descriptors such as maximum, minimum and other e-state indices (\citet{kier1999molecular}), molecular weight, Balaban's J index (\citet{balaban1982highly}) among others. The full list of descriptors used in this set is reported in Section-\ref{sec:appendix-a} of the Appendix.

We have experimented with several feature representations and observed that MolDSet works best as input features to the k-NN module (and thus as the output of the actor module) and ECFP works best as input to the $f$, $\pi$ and $Q$ networks. The results reported in this paper use only these two features. Further analysis is provided in Section-\ref{sec:appendix-a} of the Appendix.
\paragraph{Reaction Templates And Reactants}
The structures of reactants used in this study originate from the Enamine Building Block catalogue\footnote{https://enamine.net/building-blocks} Global Stock. Only $150,560$ unique (with stereo) building blocks that matched at least one reaction template as a first or second reactants were used in this study. The full list of SMILES of the building blocks can be found in the github repository of this work.

The set of reaction templates used in this study was taken from \citet{button2019automated}. Several templates were additionally manually curated to resolve occasional errors such as broken aromaticity and stereochemistry problems upon using them with RDKit \emph{RunReactants} function (version 2019.03.1). We note that only stereocenters specified in the initial reactants are kept in the products and stereocenters that would be formed during the reaction are left without specification. This is one of the limitations of using reaction templates that cannot accurately predict stereoselectivity. Since the reaction template selection is based on the reactant that will be used as the first one in the reaction, the 49 bimolecular templates were transformed into 98. For example, the ``Michael addition'' template also consists of the ``Michael addition R2''. The 15 unimolecular templates are also used, totaling 113 possible reaction templates. We additionally filter out reaction templates that have fewer than 50 available second reactants for them, resulting in a final set of 97 templates. Additional statistics and examples of reaction templates are provided in Section-\ref{sec:appendix-b} of the Appendix.

\paragraph{Datasets For QSAR Modeling}
The datasets for all three HIV targets were downloaded from ChEMBL (\citet{gaulton2017chembl}) corresponding to the following target IDs: HIV-RT - CheMBL247, HIV-Integrase - CheMBL3471, CCR5 - 274. The full datasets used for QSAR modeling are provided in the github repository. The data curation procedure is described in Section-\ref{sec:appendix-c} of the Appendix.

\subsection{Experimental Settings}
 
\paragraph{Model Setup} 
Hyper parameter tuning was performed and the following set of parameters were used in all the experiments reported in this paper. The $f$ network uses four fully connected layers with 256, 128, 128 neurons in the hidden layers. The $\pi$ network uses four fully connected layers with 256, 256, 167 neurons in the hidden layers. All the hidden layers use ReLU activation whereas the final layer uses tanh activation. Similarly, the $Q$ network also uses four fully connected layers with 256, 64, 16 neurons in the hidden layers, with ReLU activation for all the hidden layers and linear activation for the final layer. We use the Adam optimizer to train all the networks with a learning rate of 1e-4 for the $f$ and $\pi$ networks and 3e-4 for the $Q$ network. Further, we used a discount factor $\gamma=0.99$, mini batch size = 32, and soft update weight for target networks, $\tau = 0.005$. We have only used $k = 1$ (in the $k$-NN module) during both the training and inference phases of our algorithm for fair comparison.

\paragraph{Baseline Setup}
The specific baseline in this study, Random Search (RS) starts with a random initial reactant ($R^{(1)}$) followed by the selection of a random reaction template $T$, and then the random selection of a compatible reactant $R^{(2)}$. The product of the reaction is used as the $R^{(1)}$ in the next reaction. This process is repeated until the maximum allowed number of synthesis steps is reached or until the product doesn't have any reactive centers left. In this study, we define the maximum number of synthesis steps allowed in an episode to be five. The random search continues until the stop criterion such as search time or number of reactions is reached. The total number of allowed reaction steps used during random search to produce results in Table \ref{performance_table} and Table \ref{performance_table_100} is 400,000.
\subsection{Results And Analysis}

\subsubsection{Baseline comparison}

\begin{figure}
\centering
\vspace{-6mm}
\subfigure[]{\label{fig:qed_stat}\includegraphics[width=0.48\linewidth]{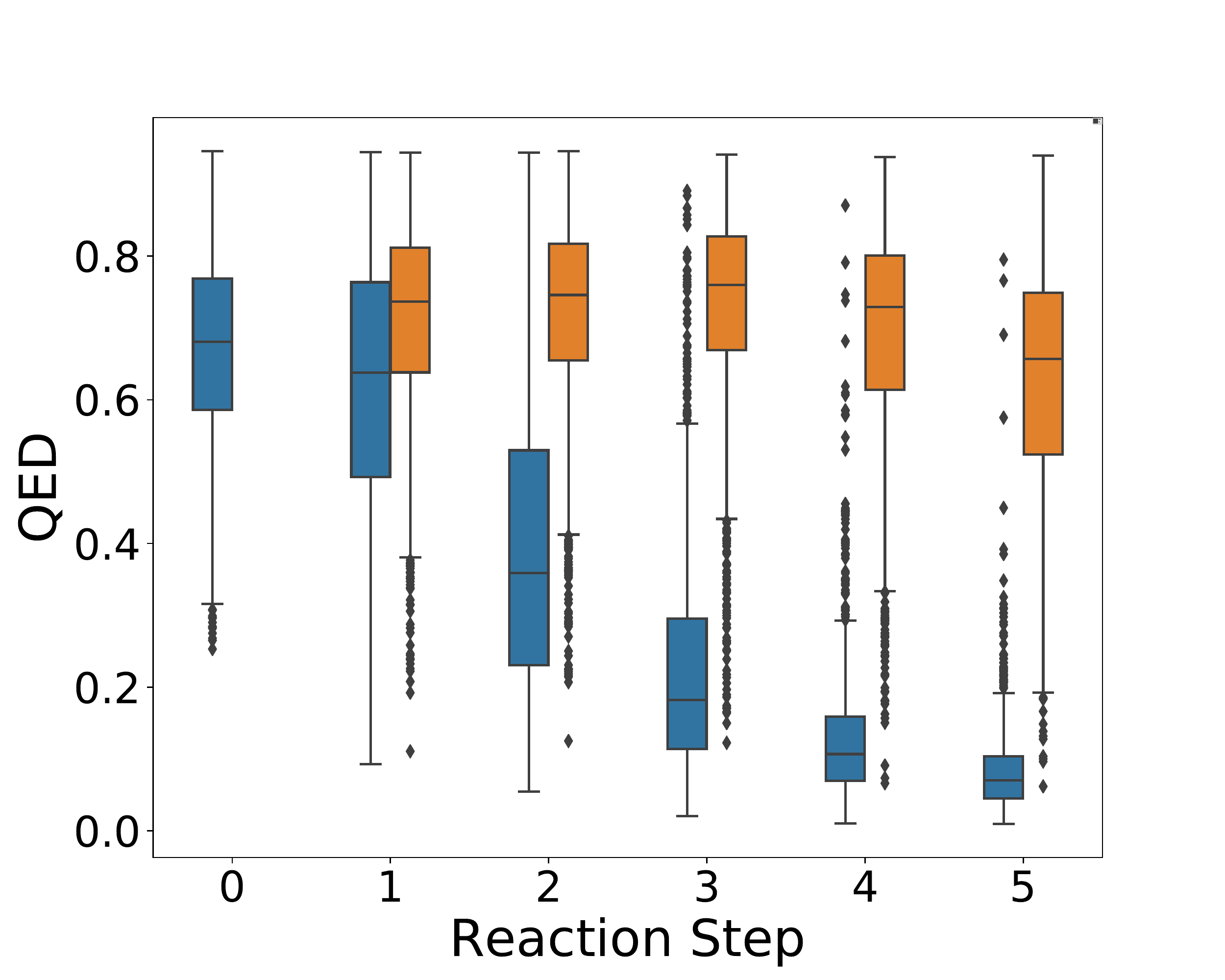}}
\subfigure[]{\label{fig:logp_stat}\includegraphics[width=0.48\linewidth]{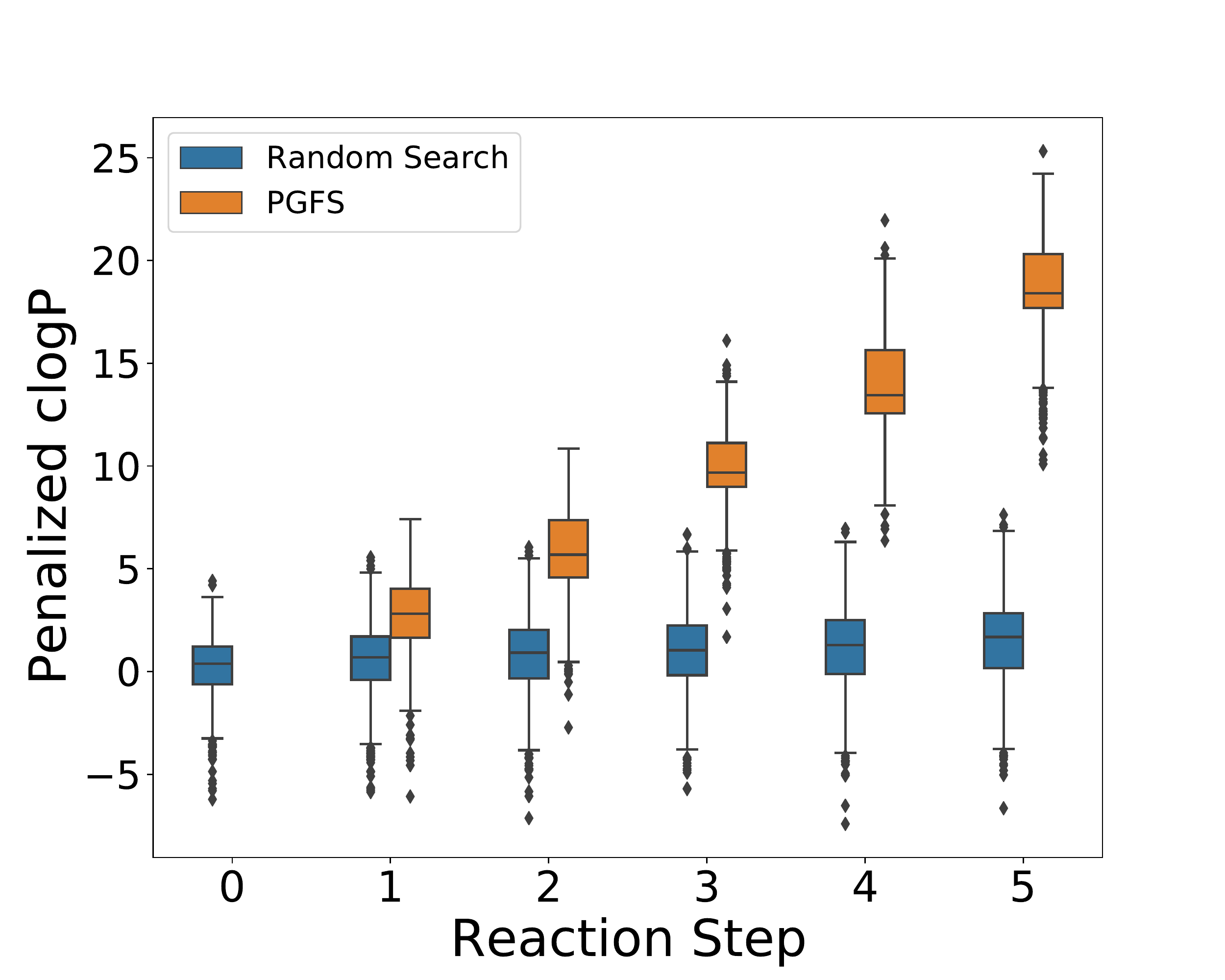}}

\hfill
\subfigure[]{\label{fig:qed_dist}\includegraphics[width=0.48\linewidth]{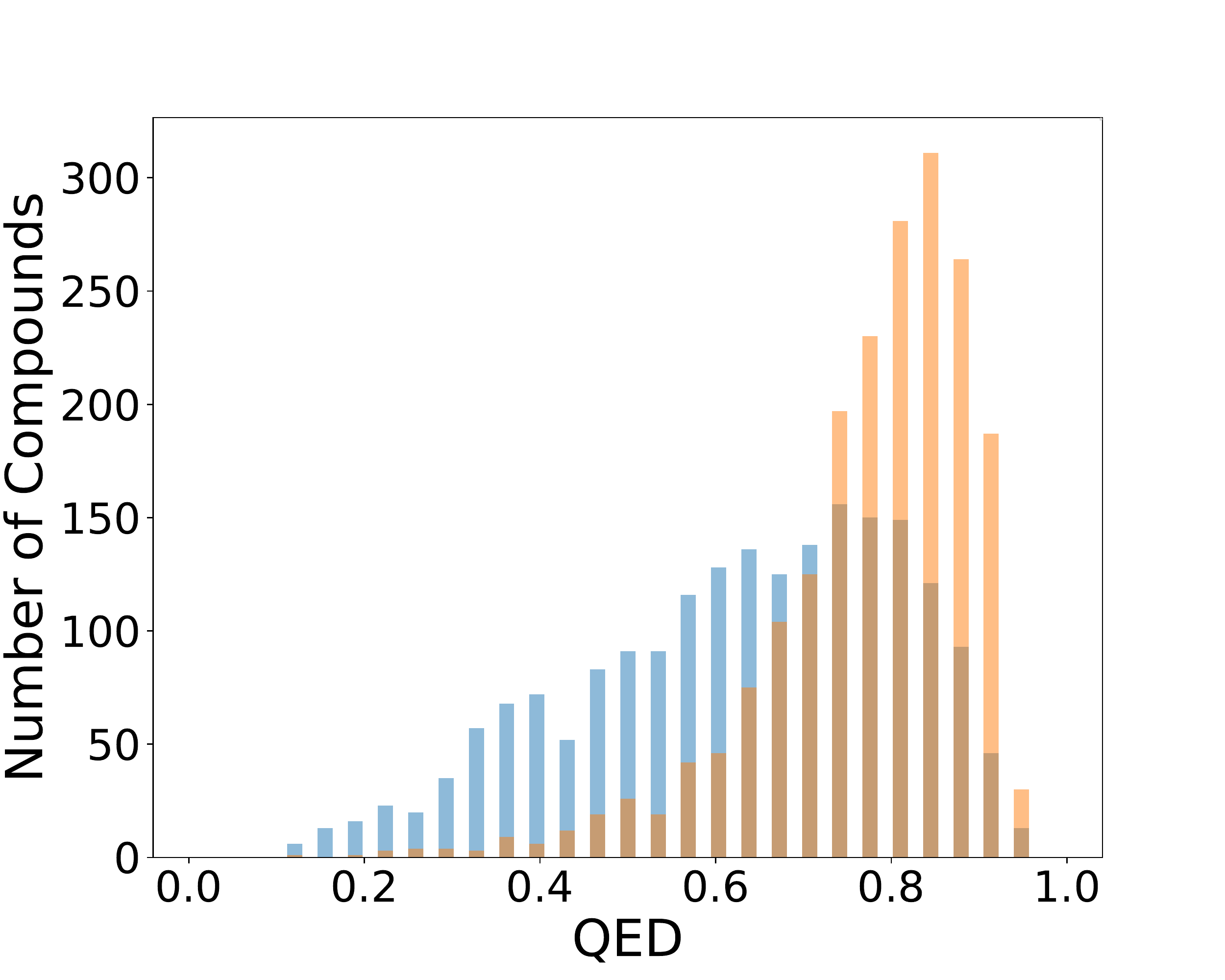}}
\subfigure[]{\label{fig:logp_dist}\includegraphics[width=0.48\linewidth]{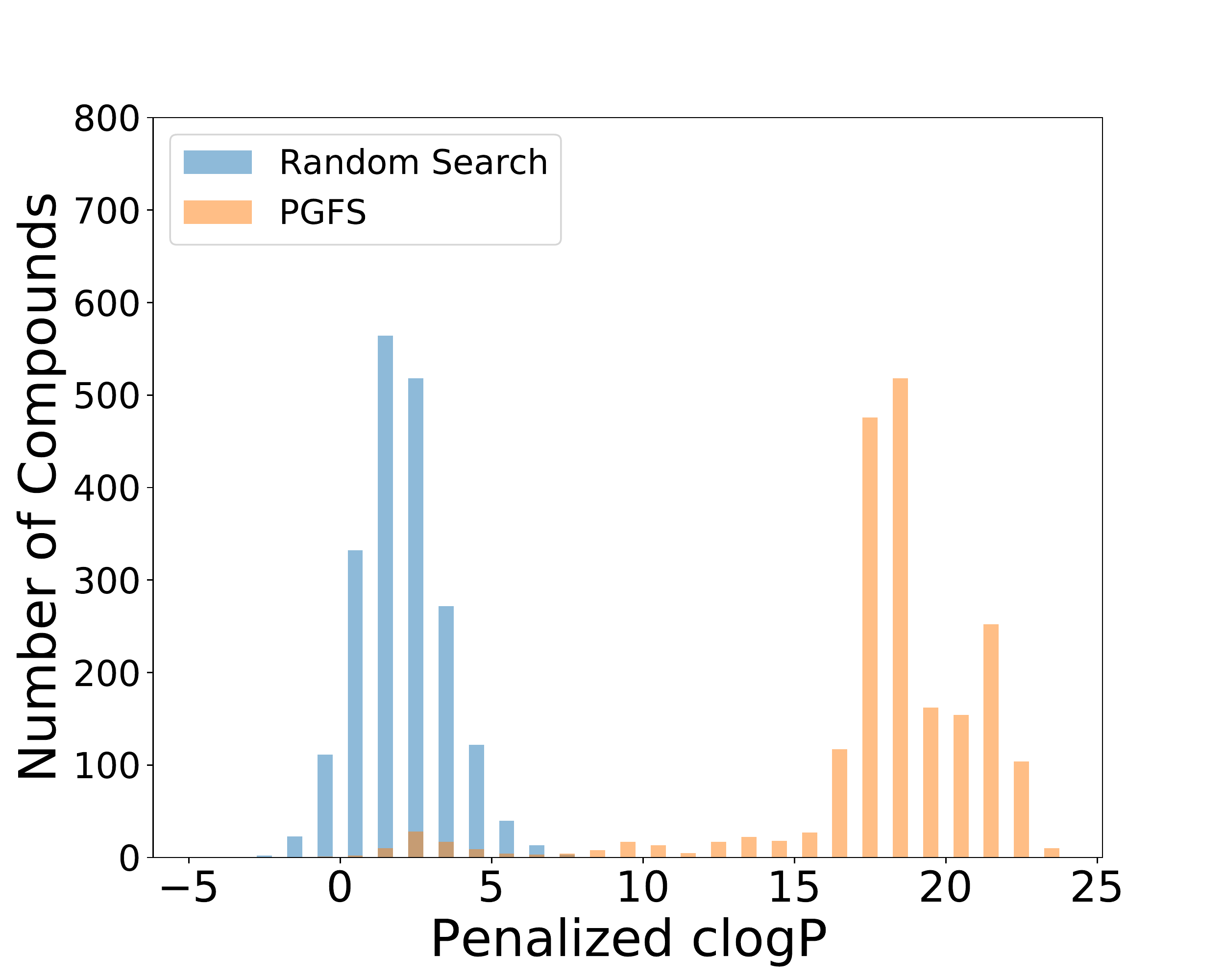}}
\caption{Performance comparison of Random Search (RS) depicted in blue vs. PGFS depicted in orange produced using the validation set of 2,000 initial reactants (R1s) using Random Search and inference run of the trained PGFS model using corresponding rewards. (a) and (b): box plots of the QED and penalized clogP scores per step of the iterative five-step virtual synthesis. The first step (Reaction Step = 0) in each box plot shows the scores of the initial reactants (R1s). (c) and (d): distributions of the maximum QED and penalized clogP scores over five-step iterations. A few outliers with penalized clogP lower than -10 with both methods were clipped out when plotting (b) and (d).}
\label{fig:comparison_logpqed}
\end{figure}

\begin{figure}[htbp]
\centering
\vspace{-8mm}
\subfigure[]{\label{fig:hiv_int_stat}\includegraphics[width=0.3\linewidth]{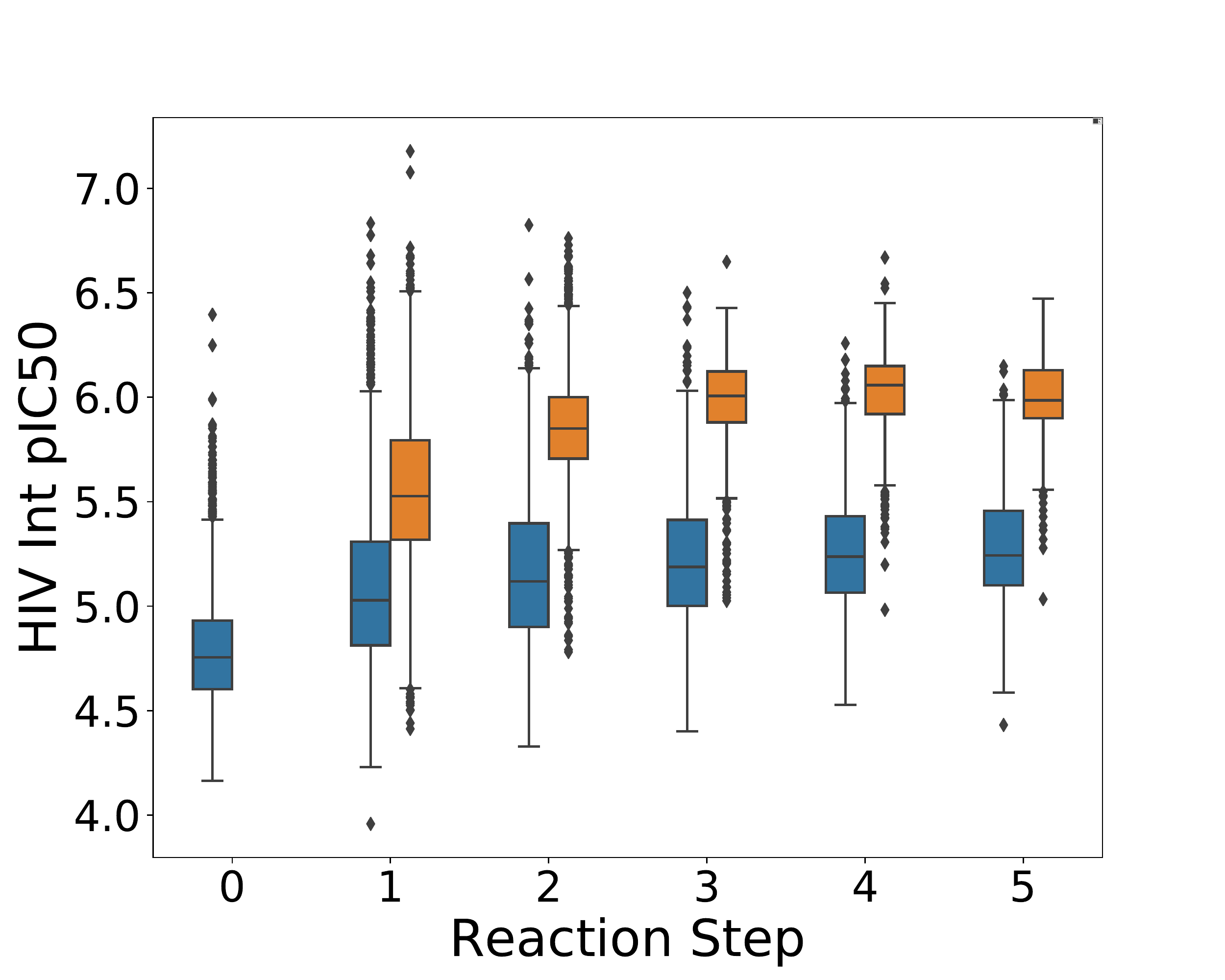}}
\subfigure[]{\label{fig:hiv_ccr5_stat}\includegraphics[width=0.3\linewidth]{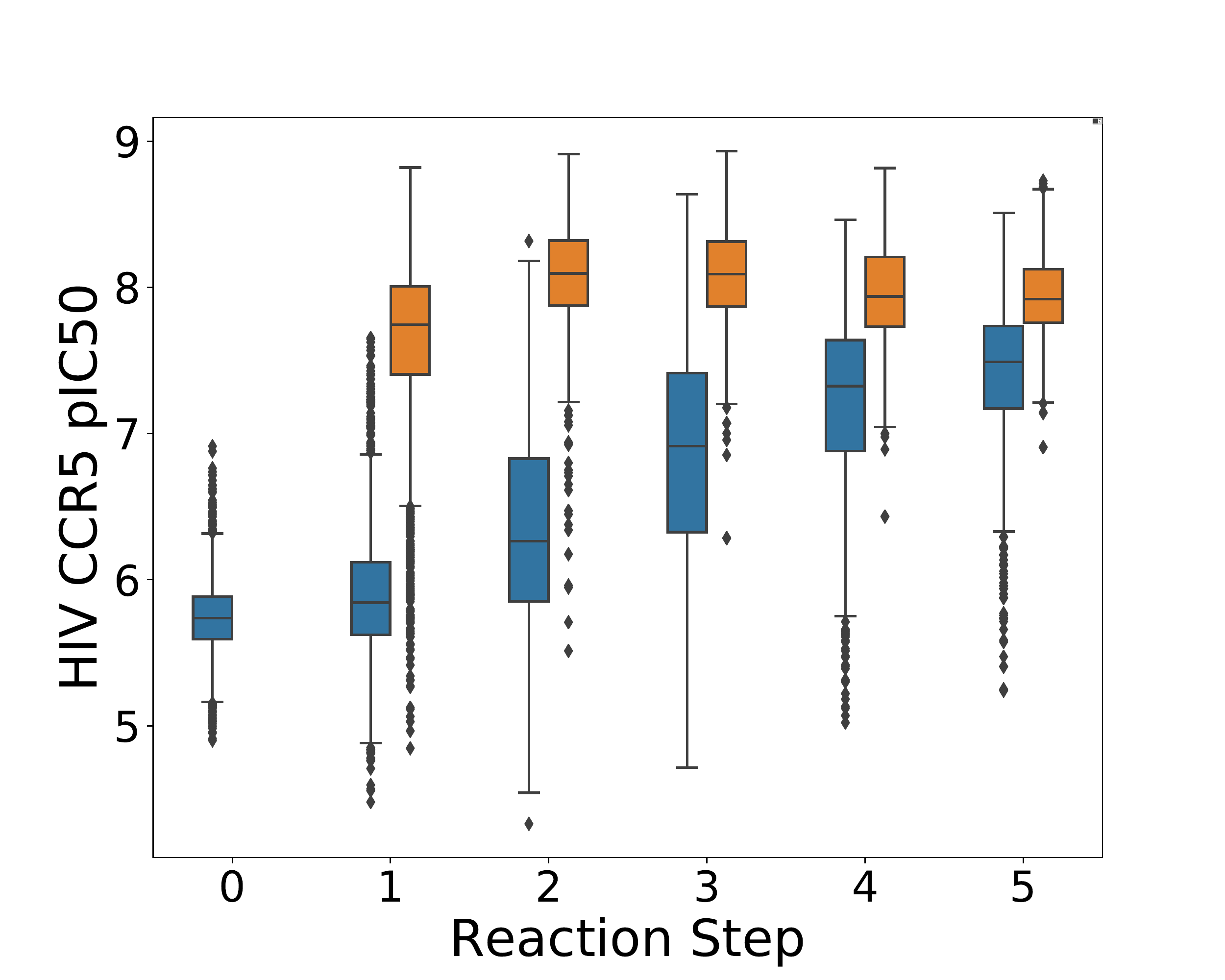}}
\subfigure[]{\label{fig:hiv_rt_stat}\includegraphics[width=0.3\linewidth]{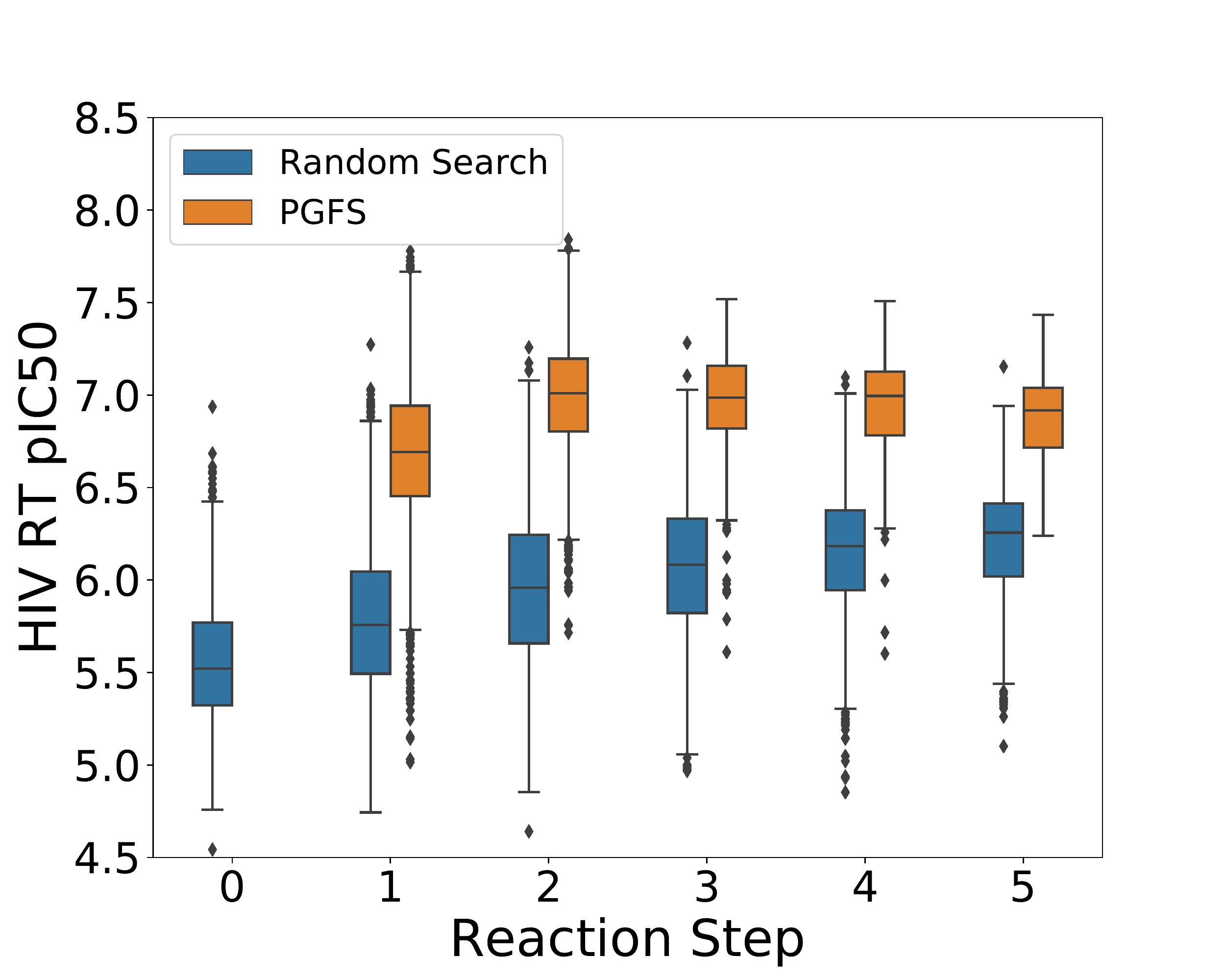}}

\hfill

\subfigure[]{\label{fig:hiv_int_dist}\includegraphics[width=0.3\linewidth]{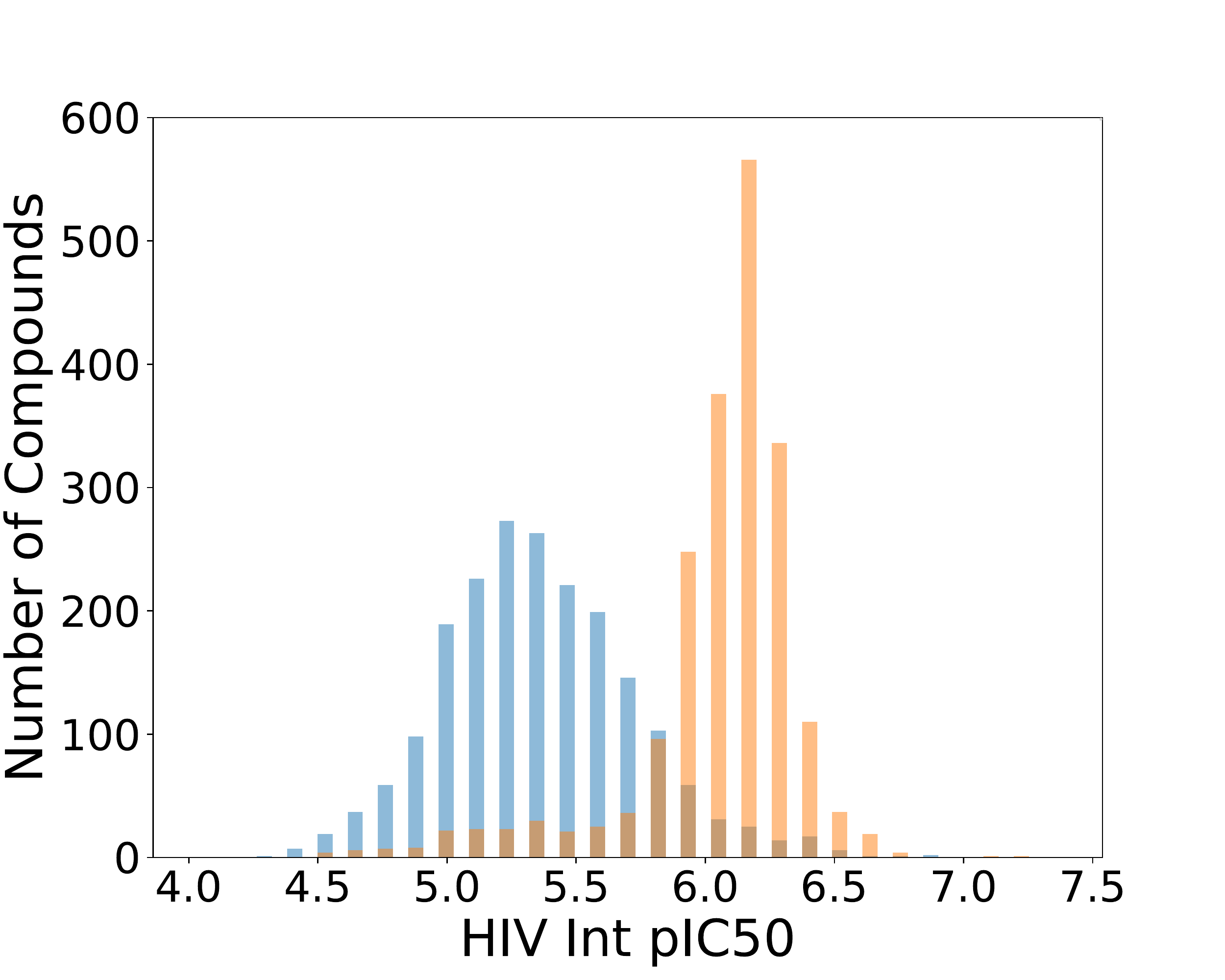}}
\subfigure[]{\label{fig:hiv_ccr5_dist}\includegraphics[width=0.3\linewidth]{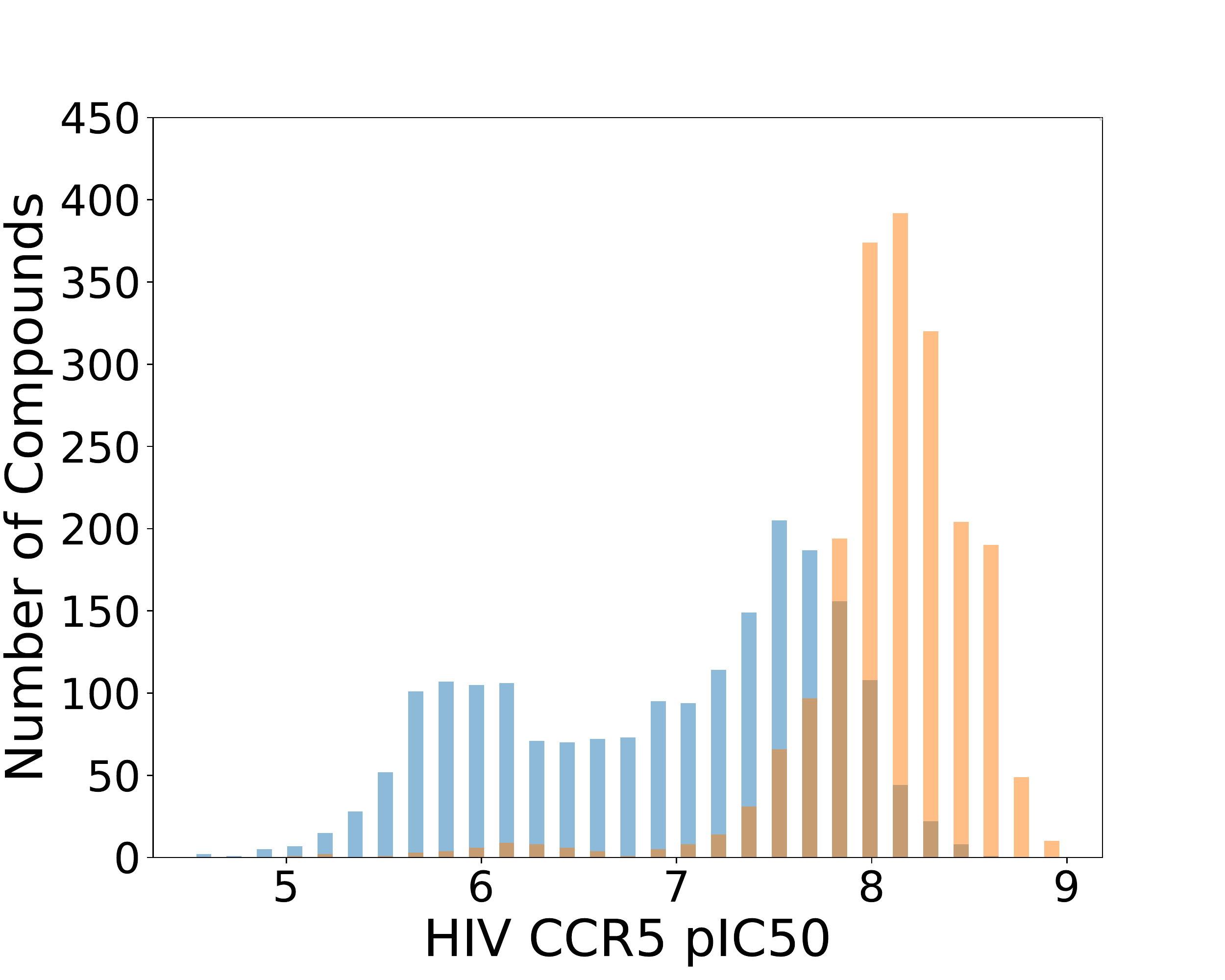}}
\subfigure[]{\label{fig:hiv_rt_dist}\includegraphics[width=0.3\linewidth]{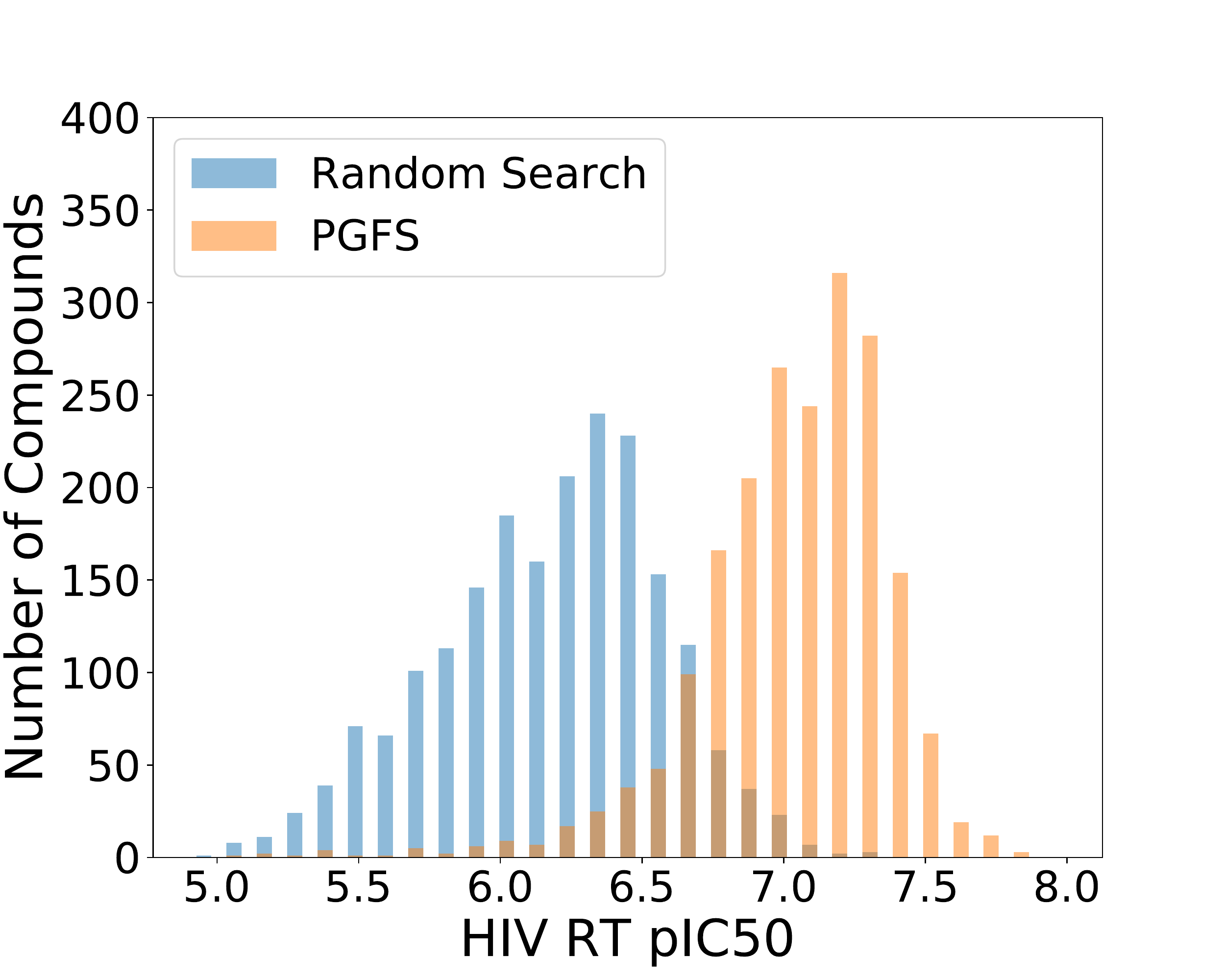}}

\hfill

\subfigure[]{\label{fig:ad_hiv_int_dist}\includegraphics[width=0.3\linewidth]{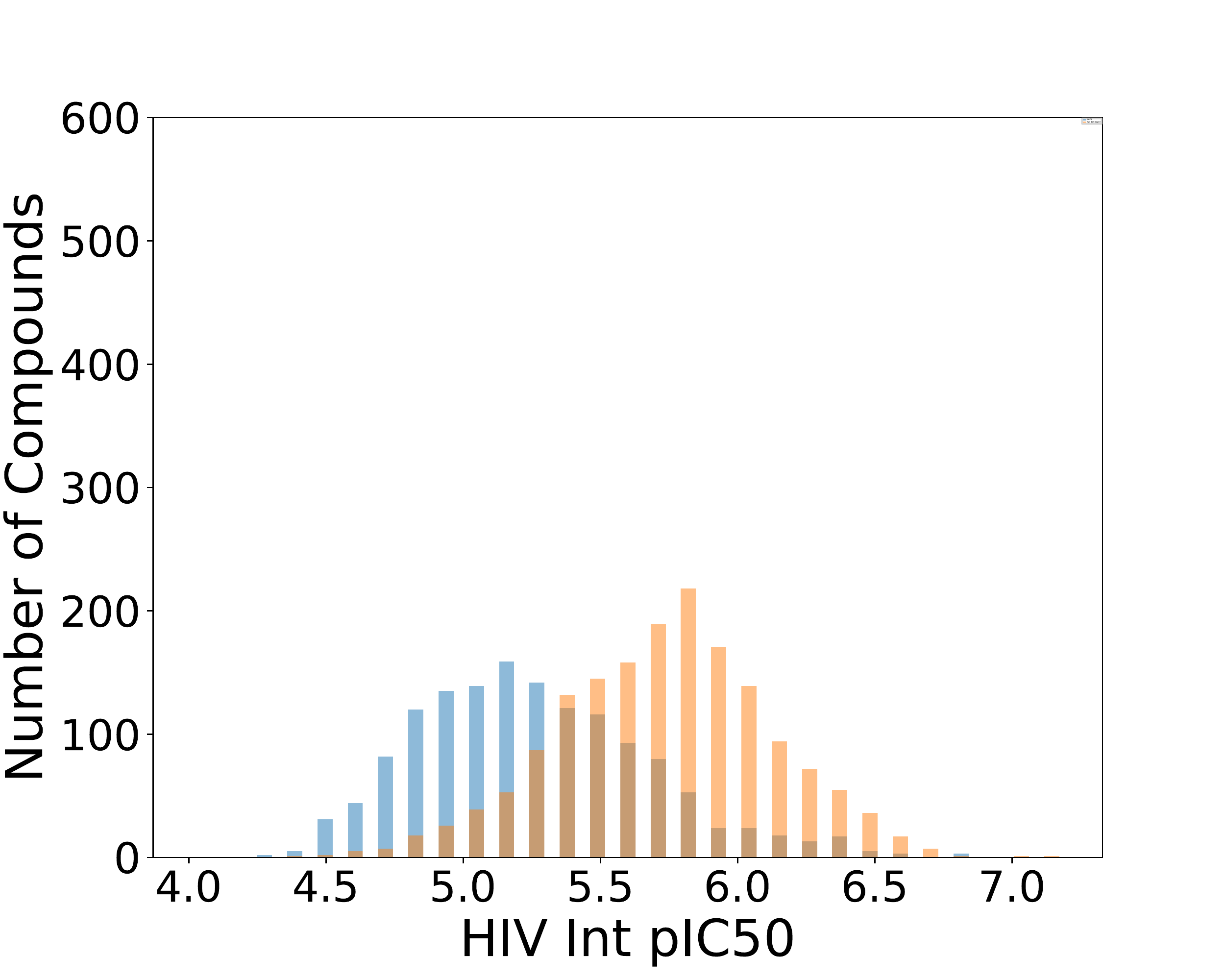}}
\subfigure[]{\label{fig:ad_hiv_ccr5_dist}\includegraphics[width=0.3\linewidth]{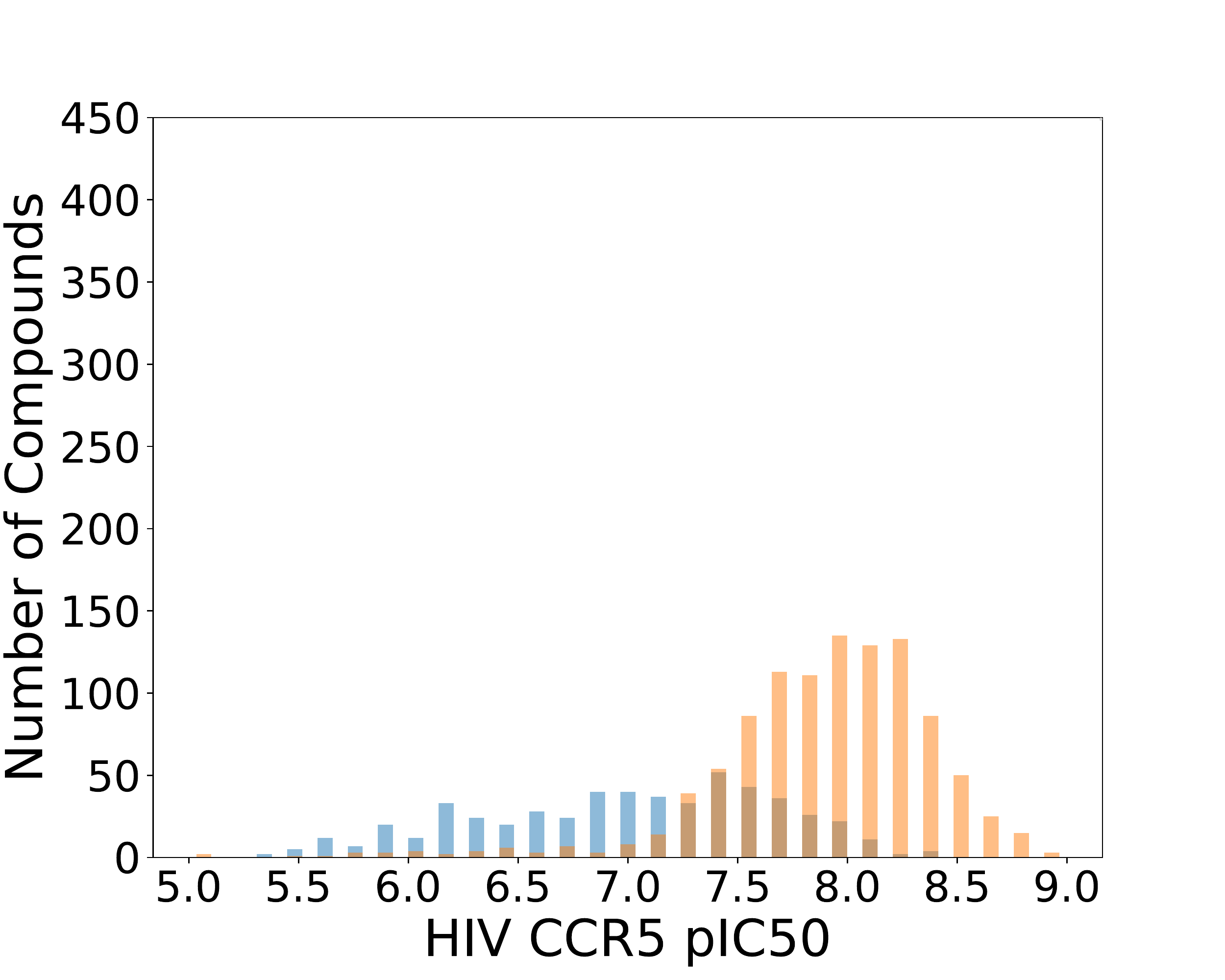}}
\subfigure[]{\label{fig:ad_hiv_rt_dist}\includegraphics[width=0.3\linewidth]{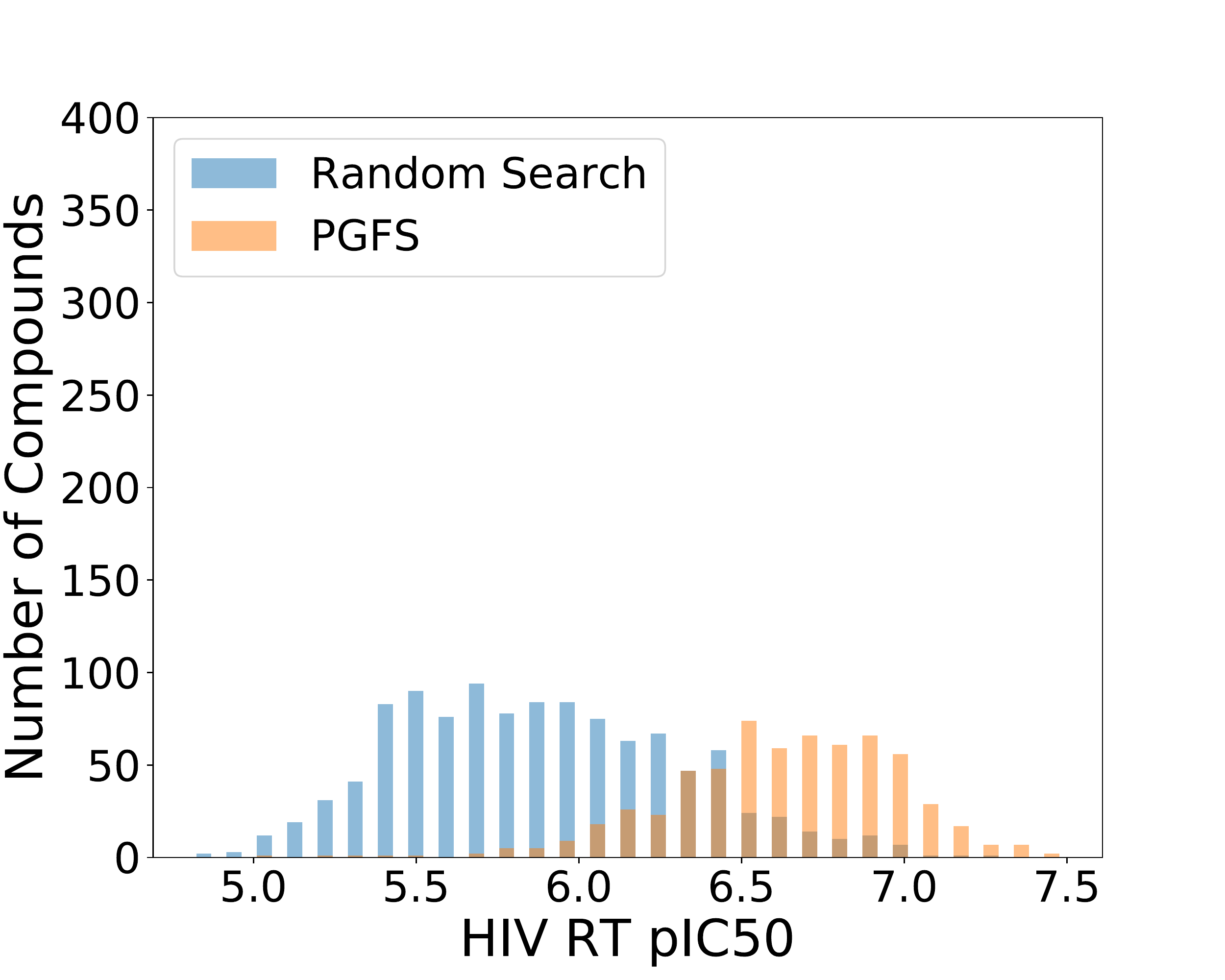}}
\caption{Performance comparison between Random Search - depicted in blue and PGFS - depicted in orange on three HIV-related QSAR-based scores. (a), (b) and (c): box plots of the corresponding QSAR-based scores per step of the iterative 5-step virtual synthesis. The first step (Reaction Step =0) in each box plot shows the scores of the initial R1s. (d), (e) and (f): distributions of the maximum QSAR-based rewards over 5-step iterations without the AD filtering. (g), (h) and (i): distributions of the maximum QSAR-based rewards over 5-step iterations after compounds that do not satisfy AD criteria of the corresponding QSAR model were filtered out from both sets.}
\label{fig:comparison_hiv}
\end{figure}
\textit{\textbf{PGFS performance on QED and penalized clogP rewards vs. Random Search(RS) - }}
The validation set constitutes randomly chosen 2,000 $R^{(1)}$s from the set of 150,560 available initial reactants. First, we carry out random search (RS) by randomly choosing reaction templates and second reactants (for bimolecular reactions) at every time step. Then, we use the trained models from PGFS (trained on QED for 200,000 time steps and on penalized clogP for 390,000 time steps) in the inference mode to predict the corresponding standard metrics, QED and penalized clogP, on this validation set. We observe that our algorithm performs significantly better than the random baseline. We can observe a clear distribution shift of each score given the same initial compounds which confirms that the training was successful. 

\textit{\textbf{PGFS performance on HIV rewards vs. Random Search(RS)}}
Next, we implement both these algorithms on HIV rewards and make a similar observation from Figure \ref{fig:comparison_hiv} that the rewards associated with the structures obtained using our method (PGFS) are substantially better than the RS method. Furthermore, we filter out compounds that do not satisfy the AD criteria of the QSAR models from both sets and still clearly observe the distribution shift towards high scoring compounds in case of PGFS in Figures \ref{fig:ad_hiv_int_dist}, \ref{fig:ad_hiv_rt_dist} and \ref{fig:ad_hiv_ccr5_dist}. PGFS was trained on HIV-CCR5 for 310,000 time steps, HIV-INT for 120,000 time steps, HIV-RT for 210,000 time steps. All these models were pre-trained on the QED task for 100,000 time steps.

\subsubsection{Quantitative performance benchmark}

\begin{table}[!ht]
\centering
\caption{Performance comparison of the maximum achieved value with different scoring functions. The reported HIV-related QSAR-based scoring functions RT, INT and CCR5 correspond to structures inside of the AD of the predictive ensemble. If a structure with maximum value is outside of the AD, its value is reported in brackets next to the maximum value inside AD. The compounds with the highest scores are presented in Appendix Section B. The QED and penalized clogP values for JT-VAE, GCPN and MSO are taken from \citet{jtvae}, \citet{gcpn},  \citet{winter2019efficient} respectively. The experiments performed to evaluate the models on HIV rewards are detailed in the appendix. Values corresponding to the initial set of building blocks are reported as ENAMINEBB.}

\def\mpmn#1#2{{\ensuremath{{#1}\pm{#2}}}}%
    \def\mpmb#1#2{{\ensuremath{{\textbf{#1}}\pm{\textbf{#2}}}}}%

\vskip 0.15in
    \resizebox{0.45\textwidth}{!}{
    \begin{tabular}{cccccc}
        \toprule
        Method & QED & clogP & RT & INT & CCR5 \\
        \midrule
        ENAMINEBB &\textbf{0.948}&5.51&7.49&6.71&8.63\\
        RS &\textbf{0.948}&8.86&7.65&7.25&8.79 (8.86)\\
        GCPN &  \textbf{0.948}& 7.98 &7.42(7.45)&6.45&8.20(8.62)\\
        JT-VAE &  0.925& 5.30 &7.58 &7.25 &8.15 (8.23) \\
        MSO &  \textbf{0.948}& 26.10 &7.76 &7.28 & 8.68 (8.77)\\
        \textbf{PGFS}  & \textbf{0.948} & \textbf{27.22}     & \textbf{7.89}      & \textbf{7.55}     &   \textbf{9.05}   \\ 
        \bottomrule
    \end{tabular}
    \label{performance_table}
     }
    
\end{table}
\begin{table}[]
\centering
\caption{Statistics of the top-100 produced molecules with highest predicted HIV scores for every method used and Enamine's building blocks. Only unique compounds were used after the stereo information was stripped to calculate the values presented in this table. *GCPN and MSO runs only produced 90 and 28 compounds inside the AD of the CCR5 QSAR model,respectively.}
\resizebox{0.65\textwidth}{!}{
\begin{tabular}{@{}ccccccc@{}}
\toprule
& \multicolumn{3}{c}{NO AD}                              & \multicolumn{3}{c}{AD}            \\ \midrule
\multicolumn{1}{c|}{Scoring}   & RT        & INT       & \multicolumn{1}{c|}{CCR5}      & RT        & INT       & CCR5      \\ \midrule
\multicolumn{1}{c|}{ENAMINEBB} & $6.87\pm{0.11}$         & $6.32\pm{0.12}$         & \multicolumn{1}{c|}{$7.10\pm{0.27}$}         & $6.87\pm{0.11}$         & $6.32\pm{0.12}$          & $6.89\pm{0.32}$         \\
\multicolumn{1}{c|}{RS}        & $7.39\pm{0.10}$ & $6.87\pm{0.13}$ & \multicolumn{1}{c|}{$8.65\pm{0.06}$} & $7.31\pm{0.11}$ & $6.87\pm{0.13}$ & $8.56\pm{0.08}$ \\
\multicolumn{1}{c|}{GCPN}      & $7.07\pm{0.10}$          & $6.18\pm{0.09}$         & \multicolumn{1}{c|}{$7.99\pm{0.12}$}         & $6.90\pm{0.13}$         & $6.16\pm{0.09}$         & $6.95\textbf{*}\pm{0.05}$         \\
\multicolumn{1}{c|}{JTVAE}      & $7.20\pm{0.12}$ & $6.75\pm{0.14}$ & \multicolumn{1}{c|}{$7.60\pm{0.16}$} & $7.20\pm{0.12}$ & $6.75\pm{0.14}$ & $7.44\pm{0.17}$ \\
\multicolumn{1}{c|}{MSO}     & $7.46\pm{0.12}$  & $6.85\pm{0.10}$  & \multicolumn{1}{c|}{$8.23\pm{0.24}$} & $7.36\pm{0.15}$ & $6.84\pm{0.10}$ & $7.92\textbf{*}\pm{0.61}$ \\
\multicolumn{1}{c|}{PGFS}       & $\textbf{7.81}\pm{0.03}$ & $\textbf{7.16}\pm{0.09}$ & \multicolumn{1}{c|}{$\textbf{8.96}\pm{0.04}$} & $\textbf{7.63}\pm{0.09}$ & $\textbf{7.15}\pm{0.08}$ & $\textbf{8.93}\pm{0.05}$ \\ \bottomrule
\end{tabular}
\label{performance_table_100}
}
\end{table}

Table \ref{performance_table} compares our proposed model performance against various models on different scoring functions (\citet{winter2019efficient, gcpn, jtvae}). Our proposed framework has produced compounds with the highest maximum scores compared to all other approaches on every defined task.
PGFS achieved a maximum QED score reported in the de novo generative design studies. However, although our system cannot just return initial building block without any reactions applied to it, we can see that a set of initial building blocks (ENAMINEBB) already contains the compounds with the highest QED of 0.948. Random search was also successful in producing a maximum QED scoring compound.  We also notice a significantly higher maximum penalized clogP value compared to the existing approaches, especially GCPN and JT-VAE. This is due to the fact that molecular size correlates with the heuristic penalized clogP score if the molecule contains hydrophobic moieties and that our method does not have restrictions (besides number of reaction steps) associated with the size of the produced structures in contrast to other methods; achievable values of the penalized clogP score  strongly depend on the number of steps the reaction-based system is allowed to take.
Thus, QED and penalized clogP scores are insufficient to compare approaches designed to be used in real drug discovery setting. However, Figure~\ref{fig:comparison_logpqed} clearly demonstrates that PGFS training procedure was successful in biasing structures of virtually synthesised compounds towards high values of these scoring functions.
In the proof of concept where the task is defined as a single-objective maximization of the predicted half maximal inhibitory concentration (pIC50) of the HIV-related targets, PGFS achieves the highest scores when compared to de novo drug design methods and random search in the maximum reward achieved (Table - \ref{performance_table}) and mean of the top-100 highest rewards (Table - \ref{performance_table_100}) comparisons, given the settings of this study.

\begin{figure}
    \centering
    \includegraphics[width=0.8\columnwidth]{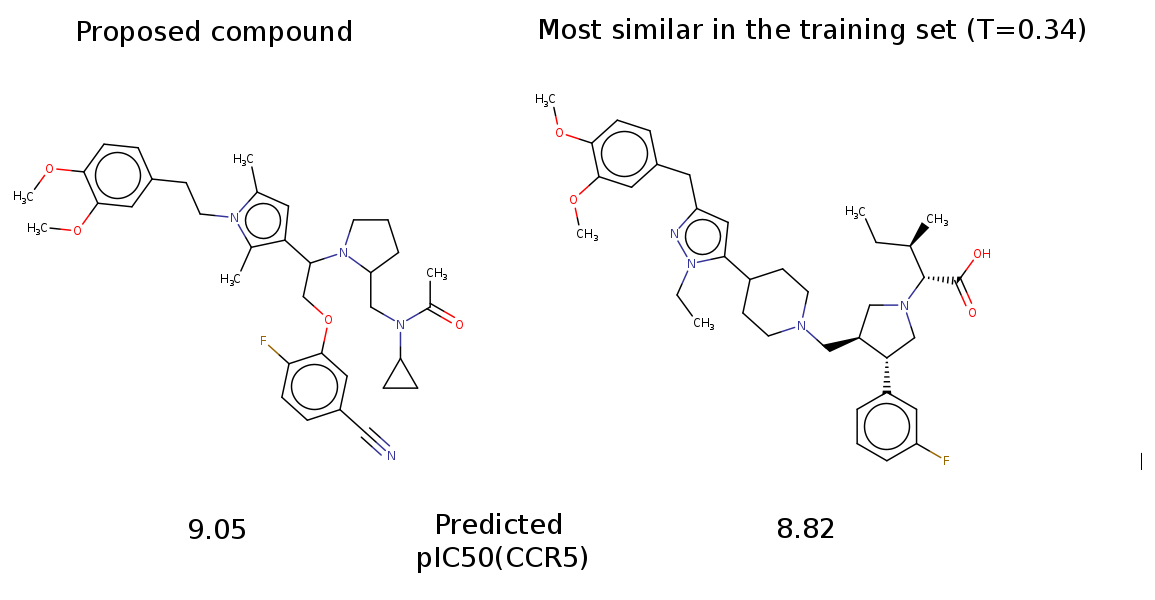}
    \vspace{-0.6cm}
    \caption{Structure of the compound proposed using PGFS with the highest predicted activity against CCR5 compared to the structure from the corresponding QSAR modeling training set with the highest Tanimoto Similarity (\citet{bajusz2015tanimoto}) using Morgan fingerprints with the radius of 2 as implemented in RDKit. The predicted pIC50 value is shown below the proposed structure and experimental pIC50 value is shown below the training set structure. 
    }
    \label{fig:ccr5_struct}
\end{figure}
\begin{figure}
    \centering
    \includegraphics[width=0.8\columnwidth]{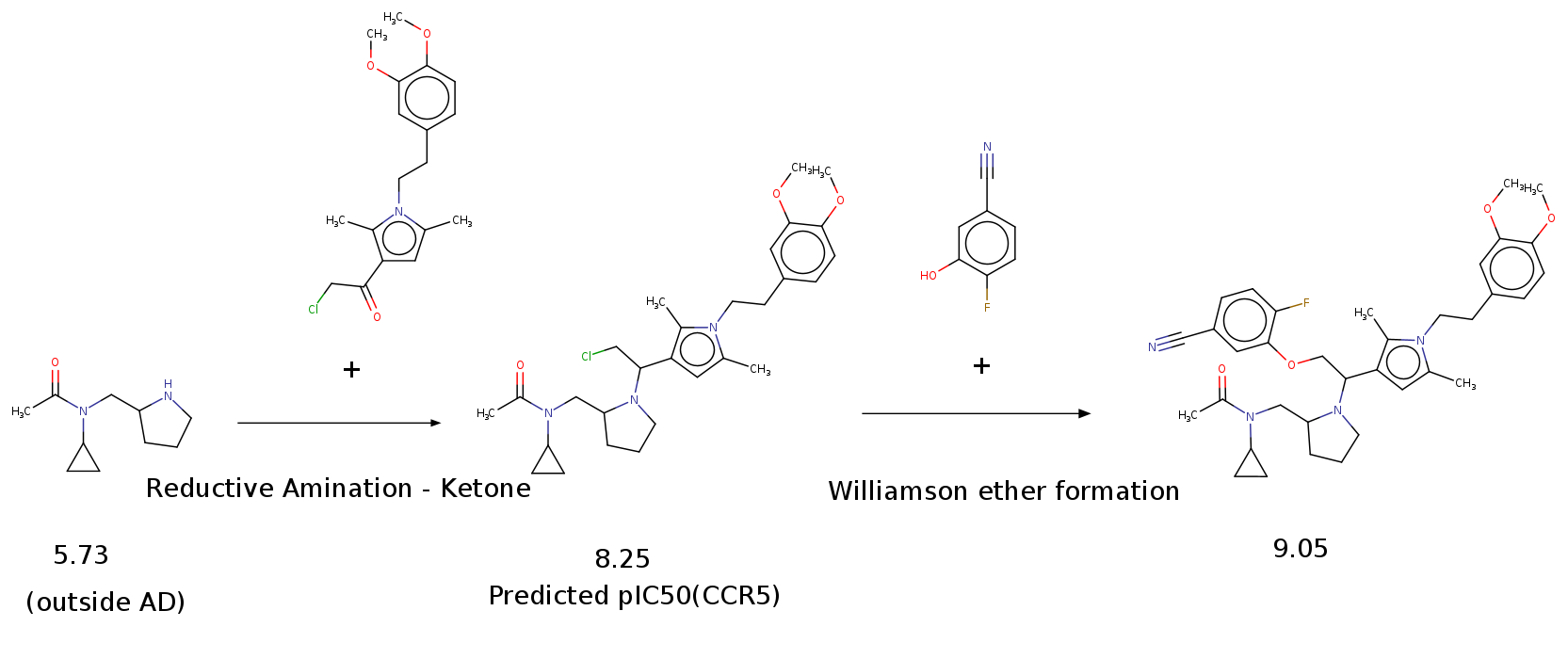}
    \vspace{-0.6cm}
    \caption{Synthesis path used by the model to produce the compound from 
    Figure~\ref{fig:ccr5_struct} with the corresponding predicted activity against CCR5 
    }
    \label{fig:ccr5_route}
\end{figure}

\paragraph{Proof-Of-Concept}
Figure \ref{fig:ccr5_struct} demonstrates one of the proposed compounds with the highest predicted inhibitory activity (pIC50) against the CCR5 HIV target. As recommended by \citet{walters2020assessing}, we also provide side by side comparison of the proposed structure with the most similar one in the training set utilized to build the QSAR model: Figure \ref{fig:ccr5_route}

\section{Conclusion and Future Work}

    In this work, we introduce the first application of RL for forward synthesis in de novo drug design, PGFS, to navigate in the space of synthesizable small molecules. We use hierarchically organized actions where the second action is computed in a continuous space that is then transformed into the best valid reactant by the environment. PGFS achieves state-of-the art performance on QED and penalized clogP tasks. We also demonstrate the superiority of our approach in an in-silico scenario that mimics the drug discovery process.
    PGFS shows stable learning across all the tasks used in this study and shows significant enrichment in high scoring generated compounds when compared with existing benchmarks. 

In future work, we propose to use a second policy gradient that solely updates the $f$ network based on the value of its corresponding critic to efficiently learn to select transformation templates (unimolecular reactions) and to stop when the expected maximum reward in an episode is attained. Furthermore, one can use any RL algorithm for continuous action space like SAC (\citet{SAC}) or a hybrid of a traditional planning and RL algorithm (\citet{deeppepper, expertiteration}). These future developments could potentially enable a better exploration of the chemical space. 


\section*{Acknowledgements}

The authors would like to thank Mohammad Amini for thoroughly reviewing the manuscript and Harry Zhao, Sitao Luan and Scott Fujimoto for useful discussions and feedback. 99andBeyond would like to thank Youssef Bennani for his valuable support and Davide Sabbadin for medicinal chemistry advice.


\bibliography{references.bib}  

\begin{appendices}{}

\section{Choice of Parameters}
\label{sec:appendix-a}

During training, for the bootstrapping phase of first 3,000 time steps, the agent randomly chooses any valid reaction template $T$ and any valid reactant $R^{(2)}$. After the bootstrapping phase, a Gaussian noise of mean 0 and standard deviation 0.1 is added to the action outputted by the $\pi$ network. No noise is added during the inference phase.

While updating the critic network, we multiply the normal random noise vector with policy noise of 0.2 and then clip it in the range -0.2 to 0.2. This clipped policy noise is added to the action at the next time step $a'$ computed by the target actor networks $f$ and $\pi$. The actor networks ($f$ and $\pi$ networks), target critic and target actor networks are updated once every two updates to the critic network. 

The representation of molecules plays a very important role in the overall performance of PGFS. We have experimented with three feature representations: ECFP, MACCS and custom features from MolDSet (RLV2), and observed that ECFP as state features and RLV2 as action features are the best representations. We used the following features in RLV2:
MaxEStateIndex, MinEStateIndex, MinAbsEStateIndex, QED, MolWt, FpDensityMorgan1, BalabanJ, PEOE-VSA10, PEOE-VSA11, PEOE-VSA6, PEOE-VSA7, PEOE-VSA8, PEOE-VSA9, SMR-VSA7, SlogP-VSA3, SlogP-VSA5, EState-VSA2, EState-VSA3, EState-VSA4, EState-VSA5, EState-VSA6, FractionCSP3, MolLogP, Kappa2, PEOE-VSA2, SMR-VSA5, SMR-VSA6, EState-VSA7, Chi4v, SMR-VSA10, SlogP-VSA4, SlogP-VSA6, EState-VSA8, EState-VSA9, VSA-EState9.

\begin{figure}

\centering
\vspace{-6mm}
\subfigure[]{\label{fig:qed_inf}\includegraphics[width=0.48\linewidth]{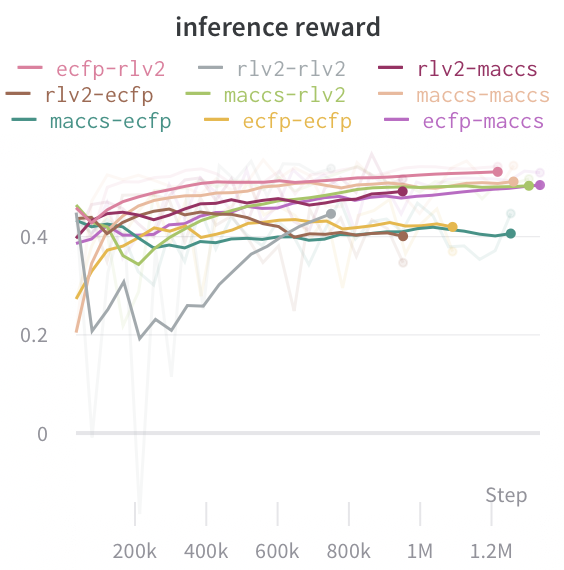}}
\subfigure[]{\label{fig:qed_f}\includegraphics[width=0.48\linewidth]{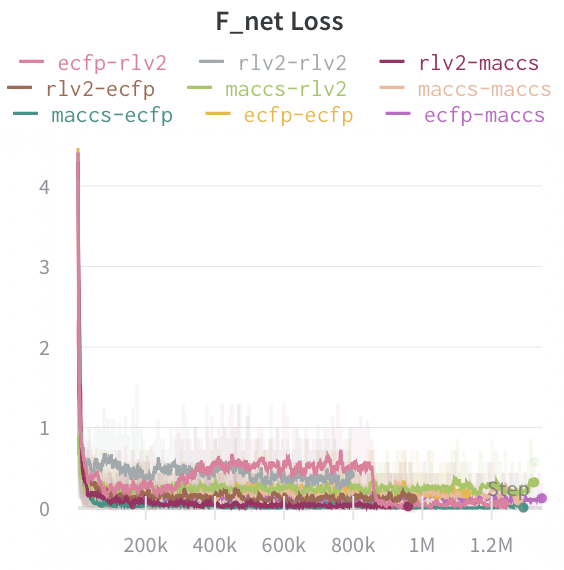}}

\hfill
\subfigure[]{\label{fig:qed_p}\includegraphics[width=0.48\linewidth]{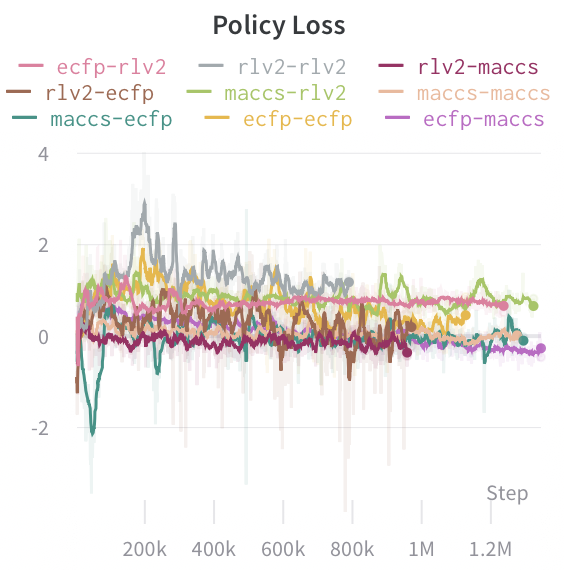}}
\subfigure[]{\label{fig:qed_v}\includegraphics[width=0.48\linewidth]{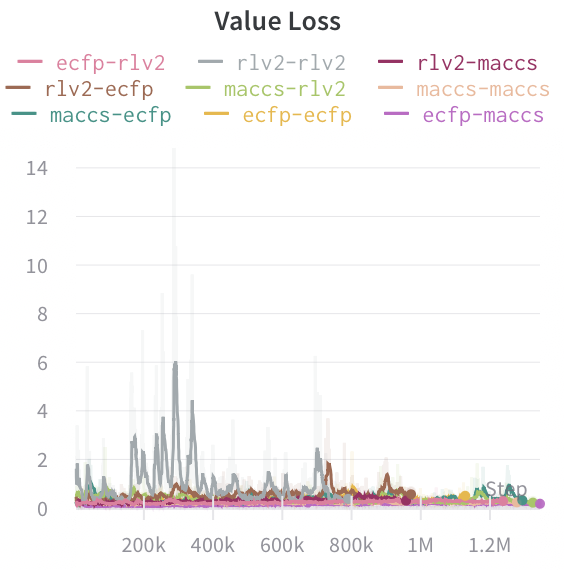}}
\caption{Plots of (a) inference reward; (b) $f$ network loss; (c) policy loss; (d) value loss; for shaped-QED reward. We can observe that ECFP as state features and RLV2 as action features (pink curve: ECFP-RLV2) performed best in terms of inference reward}
\label{fig:qed}
\end{figure}

\begin{figure}

\centering
\vspace{-6mm}
\subfigure[]{\label{fig:hiv_inf}\includegraphics[width=0.48\linewidth]{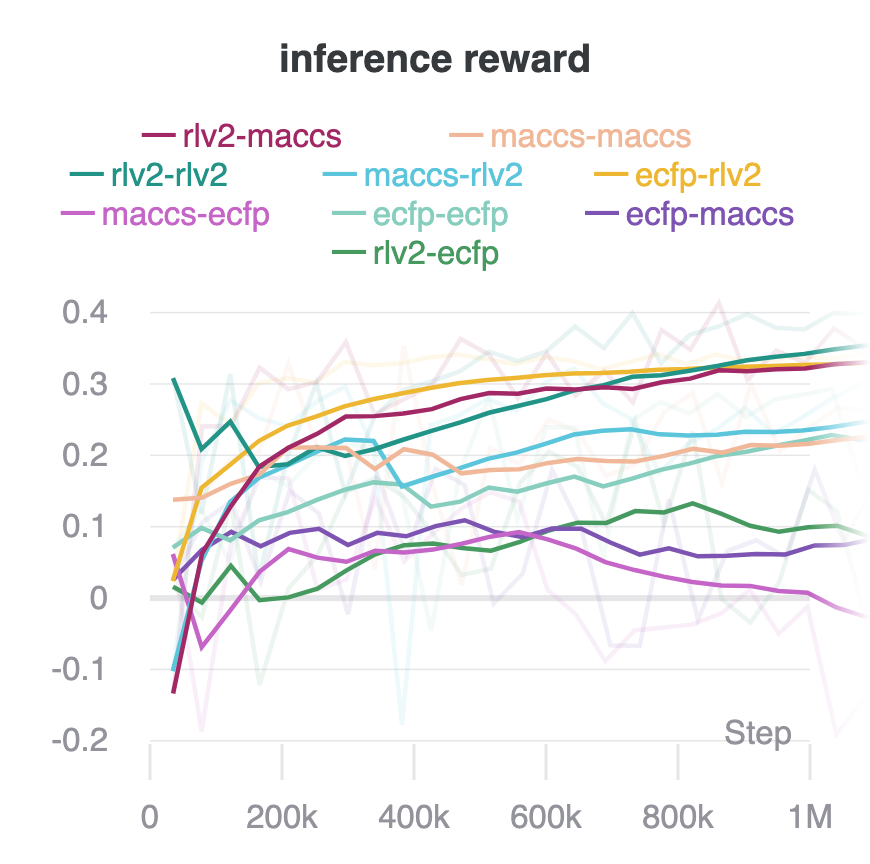}}
\subfigure[]{\label{fig:hiv_f}\includegraphics[width=0.48\linewidth]{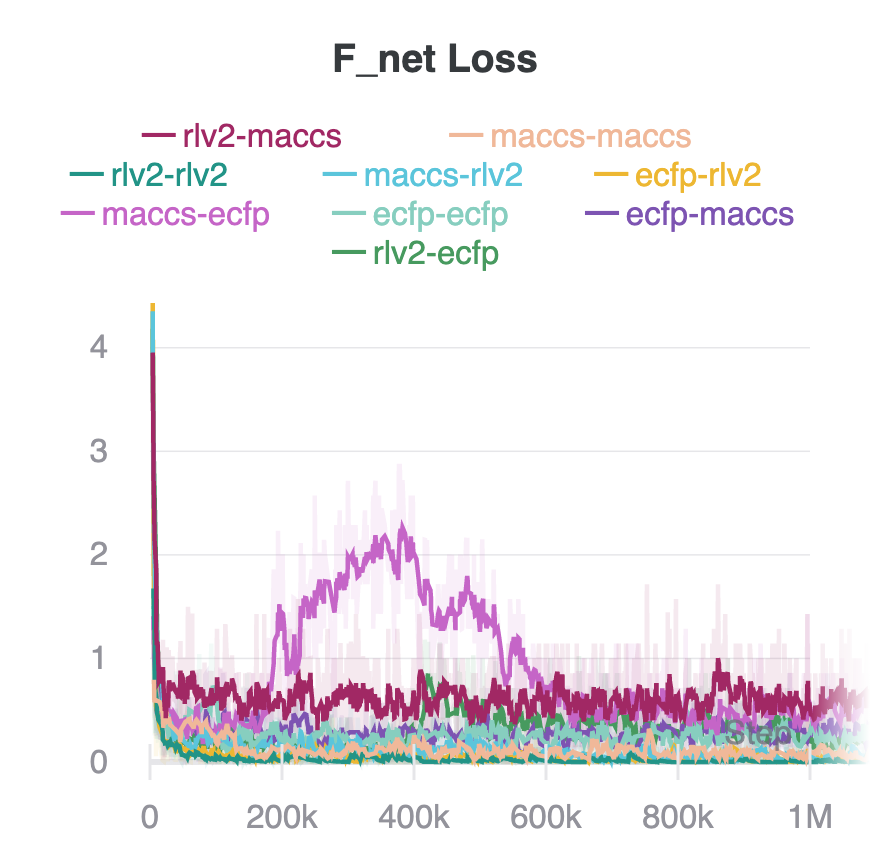}}

\hfill
\subfigure[]{\label{fig:hiv_p}\includegraphics[width=0.48\linewidth]{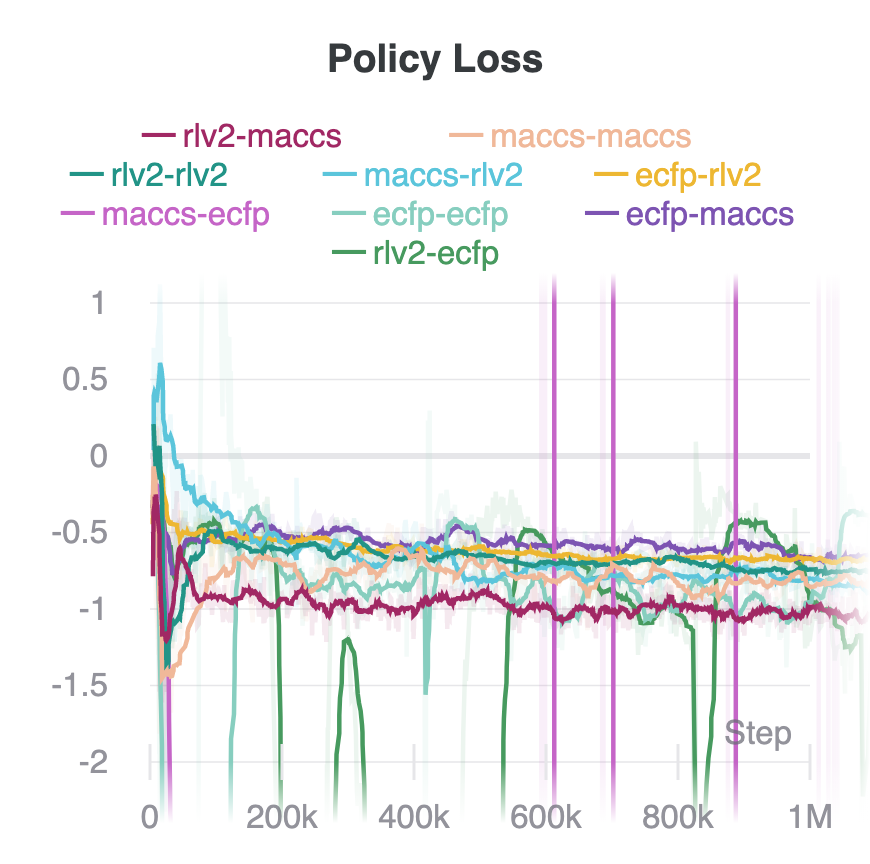}}
\subfigure[]{\label{fig:hiv_v}\includegraphics[width=0.48\linewidth]{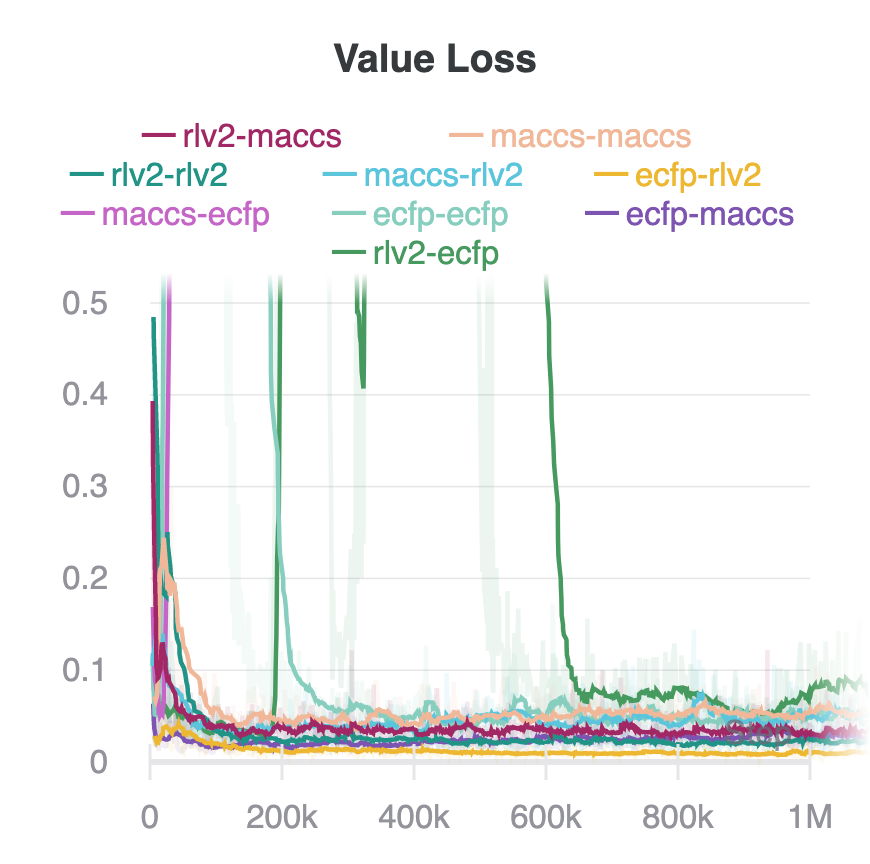}}
\caption{Plots of (a) inference reward; (b) $f$ network loss; (c) policy loss; (d) value loss; for shaped-hiv-ccr5 reward. We can observe that ECFP as state features and RLV2 as action features (yellow curve: ECFP-RLV2) performed best in terms of inference reward}
\label{fig:hiv}
\end{figure}

Figures \ref{fig:qed},\ref{fig:hiv} show the $f$ network loss, policy loss, value loss and the average (inference) reward on a set of randomly chosen 100 initial reactants as the training progresses; for all possible combinations of state and action feature representations; for shaped-QED and shaped-HIV-CCR5 rewards respectively.

\section{Examples of Proposed Compounds And Their Synthetic Paths, Additional Information On Reaction Templates And Building Blocks }
\label{sec:appendix-b}

\begin{figure}
    \centering
    \includegraphics[width=1.0\columnwidth]{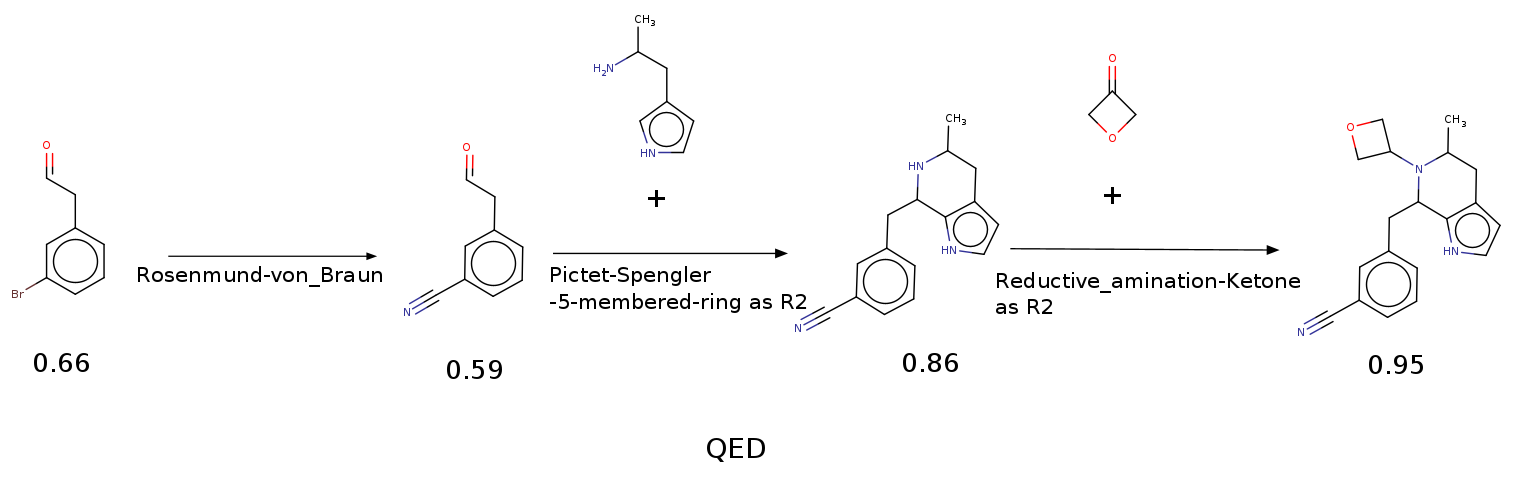}
    \vspace{-0.6cm}
    \caption{Proposed synthetic route and structure of the compound with the highest score (from the Table 2) obtained during the PGFS training using the QED score as a reward.
    }
    \label{fig:qed_route}
\end{figure}

\begin{figure}
    \centering
    \includegraphics[width=1.0\columnwidth]{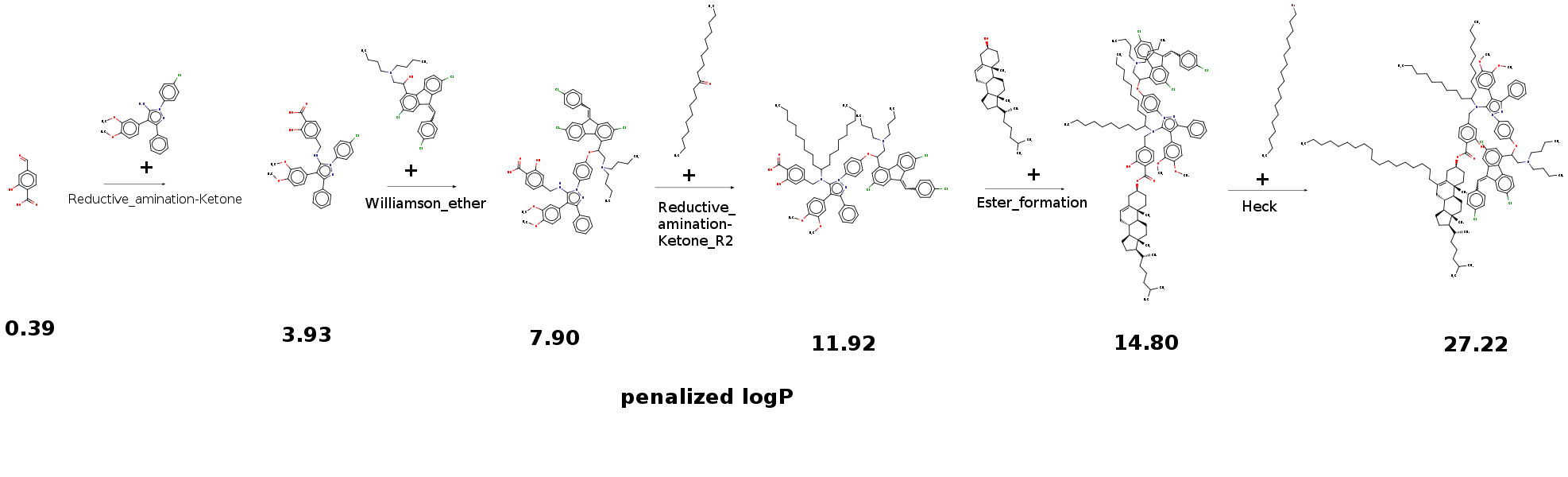}
    \vspace{-0.6cm}
    \caption{Proposed synthetic route and structure of the compound with the highest score (from the Table 2 of the manuscript) obtained during the PGFS training using the penalized clogP score as a reward.
    }
    \label{fig:logp_route}
\end{figure}

\begin{figure}
    \centering
    \includegraphics[width=1.0\columnwidth]{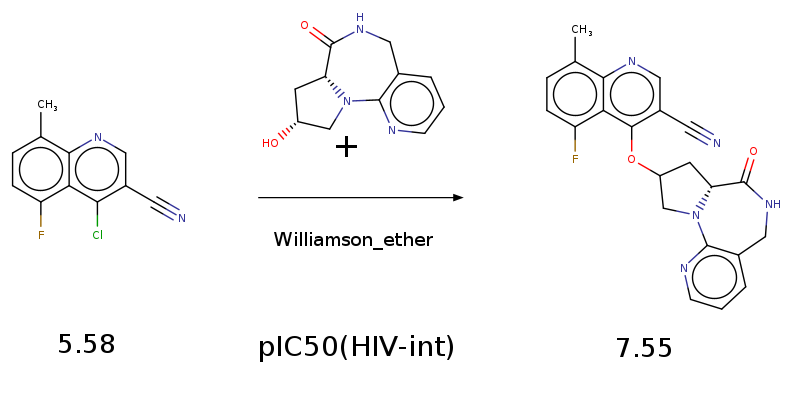}
    \vspace{-0.6cm}
    \caption{Proposed synthetic route and structure of the compound with the highest score (from the Table 2 of the manuscript) obtained during the PGFS training using predicted activity score (pIC50) against HIV-Integrase as a reward.
    }
    \label{fig:int_route}
\end{figure}

\begin{figure}
    \centering
    \includegraphics[width=1.0\columnwidth]{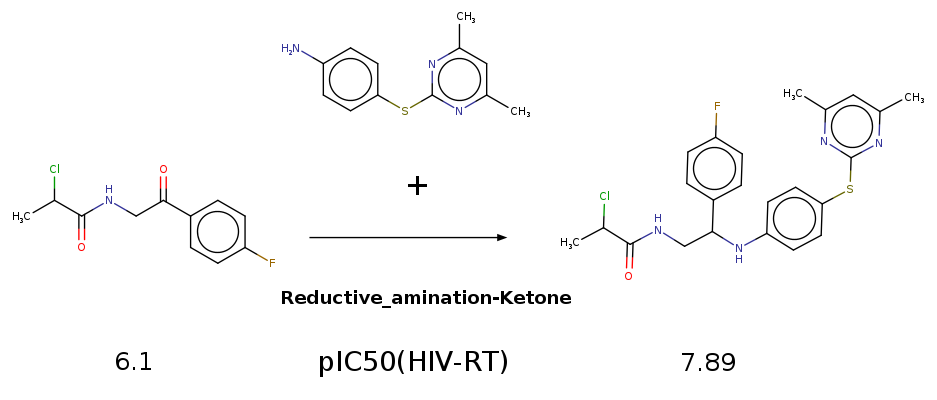}
    \vspace{-0.6cm}
    \caption{Proposed synthetic route and structure of the compound with the highest score (from the Table 2 of the manuscript) obtained during the PGFS training using predicted activity score (pIC50) against HIV-RT as a reward.
    }
    \label{fig:rt_route}
\end{figure}

Figures \ref{fig:qed_route}, \ref{fig:logp_route}, \ref{fig:int_route}, \ref{fig:rt_route} depict synthetic routes and structures of the compounds proposed by PGFS with maximum QED, penalized clogP, HIV-integrase and HIV-RT scores respectively. (The manuscript contains synthetic route and structure of the compound with maximum predicted CCR5 pIC50 and, hence, this data is not present in the Appendix to avoid duplication).

\begin{figure}[!ht]
\subfigure[]{\label{fig:templates-per-compound}\includegraphics[width=0.5\linewidth]{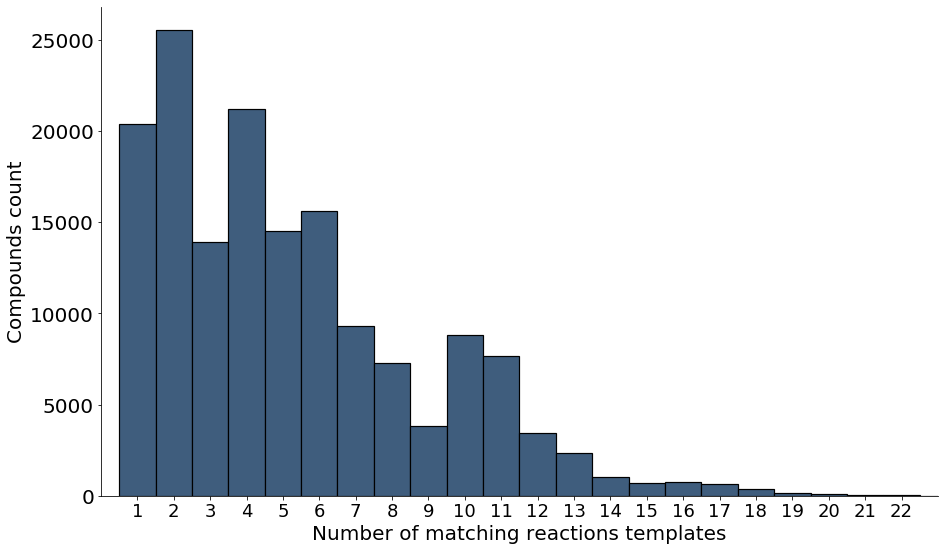}}
\subfigure[]{\label{fig:r2s-per-template}\includegraphics[width=0.5\linewidth]{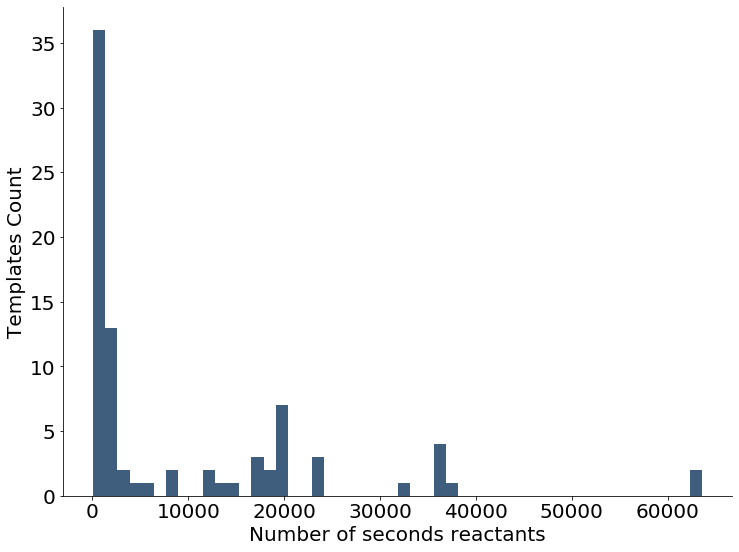}}
\caption{(a) Number of available reaction templates for compounds in the Enamine's building blocks set; (b) Number of available second reactants (R2s) from the Enamine building blocks set for reaction templates that require R2s }
\end{figure}

Figure \ref{fig:templates-per-compound} shows the distribution of available reaction templates for compounds in the Enamine building blocks dataset used in this study. Figure \ref{fig:r2s-per-template} shows how many second reactants are available for reaction templates in the set used in this study.

\section{Additional information on the QSAR-modeling}
\label{sec:appendix-c}
\subsection{Training and Cross-validation procedure }
The QSAR-based validation pipeline used in this study is similar to that reported by \citet{skalic2019target}. We have used the LightGBM implementation (\citet{ke2017lightgbm}) of the Gradient Boosting Decision Tree (GBDT) as an algorithm to train a supervised regression model on measured pIC$_{50}$ (-log$_{10}$IC$_{50}$, where the IC$_{50}$ is the concentration of a molecule that produces a half-maximum inhibitory response) values associated with 3 HIV-related targets reported in the ChEMBL database (see QSAR data collection and curation).
All the parameters of LightGBMRegressor were kept default, except for the `max-bin' and `min-split-gain' which were set to 15 and 0.0001, respectively, to decrease the complexity of the model and training times. We used 199 molecular descriptors available in RDKit (\citet{rdkit}) as an initial set of features prior to feature selection. The training procedure was performed using a five-fold cross-validation (5-CV) procedure repeated five times. To reduce the complexity of the produced models during the five-fold cross validation, after the initial set was split into training and validation sets and features were scaled using the training set of that fold random features from every pair of highly correlated features (Pearson's correlation coefficient > 0.9) were removed and a recursive feature elimination procedure was conducted using a five-fold cross-validation of the current fold training set and feature importances from LightGBMRegressor (sklearn API). The minimal required set of features with the highest adjusted determination coefficient $R_{adj}^2$ measured on 5-CV of the current fold training set were selected to build a final model. The resulting 25 models (5-CV x five times) for each HIV-target were used as an ensemble to predict pIC$_{50}$ of the produced compounds. Each model is associated with its own unique but overlapping set of selected features. The selected set of features for each model is reported in the github repository in the corresponding "list\_of\_selected\_features\_names.txt" files.  The development of multiple models for each given target makes it more difficult for the RL framework to exploit adversarial attacks on the predictor and increases the overall robustness of the predictive model.
The applicability domain (AD) (\citet{tropsha2010best}) of the ensemble of predictors was estimated using a feature space distance-based (\citet{sheridan2004similarity, sushko2010applicability}) approach. The feature sets for the AD calculation were selected as features that were used by all the 25 models in the ensemble for each target. These features are the most important for the ensemble prediction because they were selected via an independent feature selection procedure every time at every fold of the CV procedure. According to our definition of AD in this study, a compound is considered to be inside of the AD of the predictive ensemble if its average distance $ D_i$ in the normalized feature space to its K closest neighbors from the training set is equal or less than the sum of the mean of that value estimated for each point of the training set $\overline{D_t}$ and corresponding standard deviation $S$ multiplied by Z (1.5 in this study). 
$$ D_i \leq \overline{D_t} + Z*S_t$$
In this study, K used in K closest neighbors depends on the QSAR dataset and is calculated as the square root of the number of compounds in the corresponding dataset. (For CCR5 - K=41, HIV-Int - K=27, HIV-RT - K=37)

\subsection{Formulas used to calculate performance metrics in Table 1 of the manuscript }
$R^2$ - coefficient of determination is calculated as:
$$ R^{2} = \frac{ESS}{TSS} = 1 - \frac{RSS}{TSS} = 1 - \frac{\sum (e^{2}_{i}) }{\sum ( y_{i} - \bar{y})^{2} } $$
, where ${TSS}$ - Total Sum of Squares, ${ESS}$ - Explained Sum of Squares, ${RSS}$ - Residual Sum of Squares. $e^{2}_{i}$ - error term.

$R_{adj}^2$ - adjusted coefficient of determination is calculated as :
$$ R_{adj}^2 = 1 - (1 - R^2)*\frac{n - 1}{n - (p + 1)} $$,
where $R^2$ - coefficient of determination, $n$ - number of instances in the dataset, $p$ - number of features used. When calculating cross-validation performance for each separate fold $p$ is number of features used in the final model of that fold after feature selection; in the "aggregated" statistics $p$ is approximated as the average number of features used by 25 models produced by 5-CV repeated five times.

\subsection{QSAR data collection and curation}
As stated in the manuscript the datasets for QSAR modeling were downloaded from ChEMBL25 database. Only structures with available pChEMBL values corresponding to pIC50 were selected. Potential duplicates were removed. 
Standardization of chemical structures and removal of salts was done using MolVS (\citet{MolVS}). After standardizing entries with the same Canonical SMILES and different pChEMBL (pIC50) values were treated as different measurements of the same compounds, meaning that corresponding pChEMBL(pIC50) values were averaged and only unique compounds were used in further modeling. If a standard deviation of different pIC50 measurements of the same compound exceeded 1.0, the compound was discarded from the dataset. Only compounds with molecular weight (MW) between 100 and 700 were used. Outliers with Z-scores higher than 3.0 calculated using the feature values `QED', `MolWt, `FpDensityMorgan2', `BalabanJ', `MolLogP' were discarded as well.
In total, 1,719 compounds with unique Canonical SMILES for CCR5, 775 for HIV ingrase and 1,392 for HIV-RT passed all pre-processing and curation procedures. The final curated datasets used for QSAR modeling are reported in the github repository of this project.

\begin{figure}[!t]
\vspace{-8mm}
\subfigure[]{\label{fig:hiv_ccr5_perf}\includegraphics[width=0.33\linewidth]{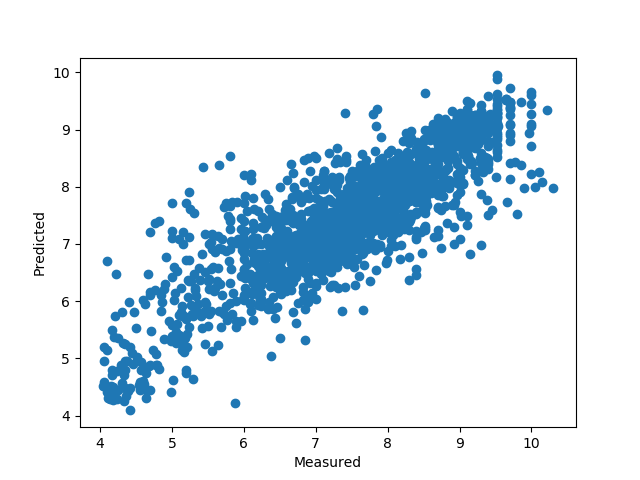}}
\subfigure[]{\label{fig:hiv_int_perf}\includegraphics[width=0.33\linewidth]{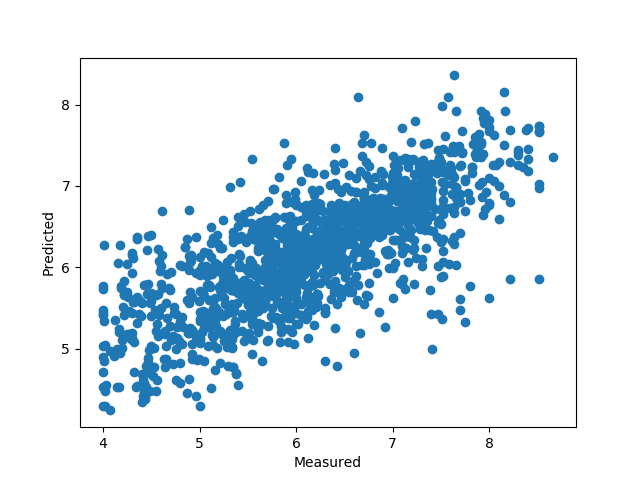}}
\subfigure[]{\label{fig:hiv_rt_perf}\includegraphics[width=0.33\linewidth]{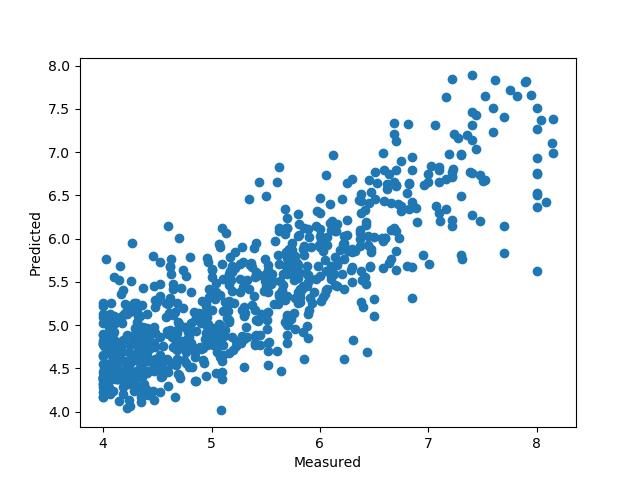}}
\caption{Cross-validation performance of QSAR modeling. Predicted by QSAR models vs. Measured plots of pIC50 values corresponding to HIV targets, from left to right: CCR5(a), HIV-RT(b), HIV-Integrase(c). Each compound instance was predicted as average prediction of only 5 models that were built while the instance was not present in the training set during the 5-fold cross validation repeated 5 times.}
\end{figure}

\section{Experimental setup for GCPN, JTVAE and MSO}
\label{sec:appenix-d}
The experiments were performed based on the experimental setting detailed in the respective publications: (\citet{you2018graph}), JT-vae (\citet{jtvae}) and MSO (\citet{winter2019efficient}). The details of which are given below.
\begin{enumerate}
    \item GCPN: The setup was run on a 32-core machine with a wall-clock time of 30 hrs, with the hyper parameters, training code and the dataset provided by the authors in their publicly available repository.
    \item JT-VAE: The setup uses the pre-trained weights provided by the authors. For the bayesian optimization (BO), as suggested by the authors, we train a sparse Gaussian process with 500 inducing points to predict properties (HIV rewards) of molecules. Then, we use five BO iterations along with expected improvement to get the new latent vectors. 50 latent vectors are proposed in each run, we obtain the molecules corresponding to them using the decoder and add them to the training set for the following iteration. Ten such independent runs are performed and the results are combined. 
    \item MSO: We use the setup corresponding to the GuacaMol (\citet{brown2019guacamol}) benchmark as provided by the authors in their work. More precisely, for each HIV reward function, a particle swarm with 200 particles was run for 250 iterations and 40 restarts. 
    
\end{enumerate}{}
\end{appendices}

\end{document}